\documentclass[5p,times,twocolumn]{elsarticle}
\usepackage{lineno,hyperref}
\modulolinenumbers[5]

\usepackage{arydshln}
\usepackage[american]{babel}
\usepackage{graphicx}
\graphicspath{{figure/}{photo/}}
\usepackage[caption=false]{subfig}
\usepackage{amsmath,amssymb}
\usepackage[bold]{hhtensor}
\usepackage{multirow} 
\usepackage{color} 
\usepackage{graphics}

\newcommand{\squishlist}{
 \begin{list}{$\bullet$}
  { \setlength{\itemsep}{0pt}
     \setlength{\parsep}{1pt}
     \setlength{\topsep}{1pt}
     \setlength{\partopsep}{0pt}
     \setlength{\leftmargin}{1.5em}
     \setlength{\labelwidth}{1em}
     \setlength{\labelsep}{0.5em} } }
\newcommand{\squishend}{
  \end{list}  }

\DeclareRobustCommand\onedot{\@onedot}
\def\@onedot{.}
\def\eg{\emph{e.g}\onedot} 
\def\ie{\emph{i.e}\onedot}

\def\etc{\emph{etc}\onedot}

\def\etal{\emph{et al}\onedot}

\DeclareMathOperator*{\erf}{erf}
\DeclareMathOperator*{\sign}{sign}
\newsavebox\CBox
\def\textBF#1{\sbox\CBox{#1}\resizebox{\wd\CBox}{\ht\CBox}{\textbf{#1}}}

\journal{Pattern Recognition}

\begin{document}

\begin{frontmatter}

\title{Learning Expectation of Label Distribution for Facial Age and Attractiveness Estimation}

\author[youtu]{Bin-Bin Gao}

\author[nju]{Xin-Xin Liu}

\author[youtu]{Hong-Yu Zhou}

\author[nju]{Jianxin Wu\corref{mycorrespondingauthor}}
\cortext[mycorrespondingauthor]{Corresponding author}
\ead{wujx2001@gmail.com}

\author[esu]{Xin Geng}

\address[youtu]{YouTu Lab, Tencent, Shenzhen 518075, China}
\address[nju]{National Key Laboratory for Novel Software Technology, Nanjing University, Nanjing 210023, China.}
\address[esu]{MOE Key Laboratory of Computer Network and Information Integration, Southeast University, Nanjing 211189, China}

\begin{abstract}
Facial attributes (\eg, age and attractiveness) estimation performance has been greatly improved by using convolutional neural networks. However, existing methods have an inconsistency between the training objectives and the evaluation metric, so they may be suboptimal. In addition, these methods always adopt image classification or face recognition models with a large amount of parameters, which carry expensive computation cost and storage overhead. In this paper, we firstly analyze the essential relationship between two state-of-the-art methods (Ranking-CNN and DLDL) and show that the Ranking method is in fact learning label distribution implicitly. This result thus firstly unifies two existing popular state-of-the-art methods into the DLDL framework. Second, in order to alleviate the inconsistency and reduce resource consumption, we design a lightweight network architecture and propose a unified framework which can jointly learn facial attribute distribution and regress attribute value. The effectiveness of our approach has been demonstrated on both facial age and attractiveness estimation tasks. Our method achieves new state-of-the-art results using the single model with 36$\times$ fewer parameters and 3$\times$ faster inference speed on facial age/attractiveness estimation. Moreover, our method can achieve comparable results as the state-of-the-art even though the number of parameters is further reduced to 0.9M (3.8MB disk storage).
\end{abstract}


\end{frontmatter}

\section{Introduction}
The human face contains a lot of important information related to individual characteristics, such as identity, expression, age, attractiveness and gender. Such information has been widely applied in real-world applications such as video surveillance, customer profiling, human-computer interaction and person identification. Among these tasks, developing automatic age and attractiveness estimation methods has become an attractive yet challenging topic in recent years. 

\begin{figure}
 \centering
 \subfloat[]
 {\includegraphics[width= 0.45\columnwidth]{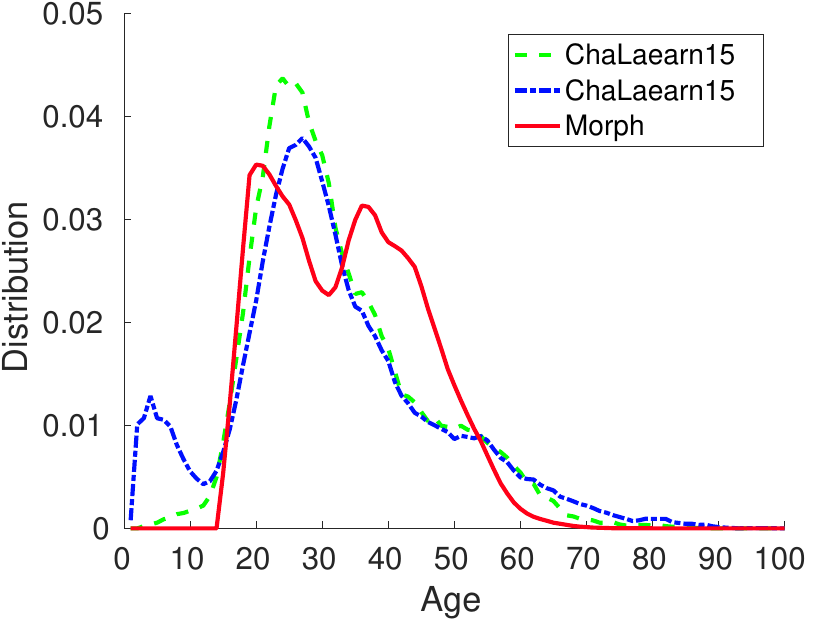}\label{fig:agedis}}
 \subfloat[]
 {\includegraphics[width= 0.45\columnwidth]{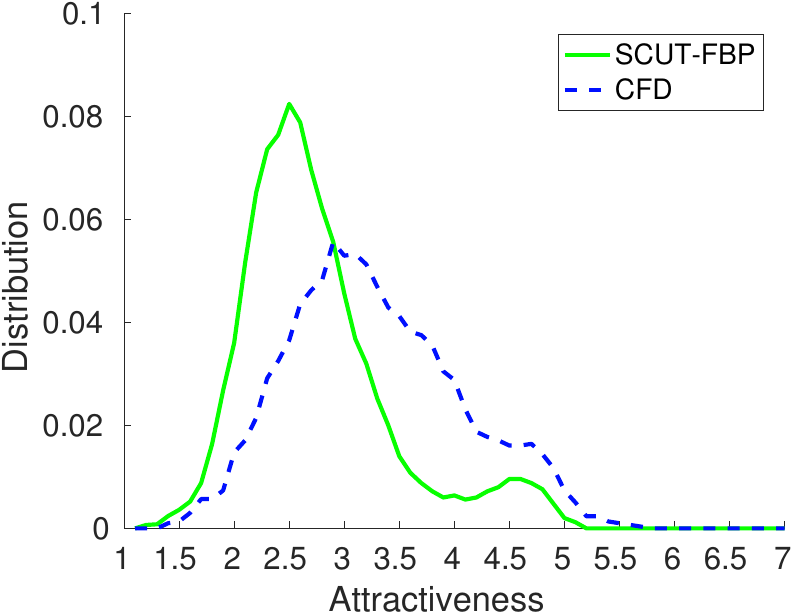}\label{fig:attdis}}
\caption{Illustration of facial attributes' distributions. (a) illustrates distributions of three facial age datasets, and (b) illustrates distributions of two facial attractiveness datasets~(Best viewed in color).} \label{fig:datadis}
\vspace{-10pt}
\end{figure}

Why is it a challenging task to find age/attractiveness from facial images? First, compared with image classification~\cite{deng2009imagenet} or face recognition~\cite{parkhi2015deep,guo2016ms,shakeel2019deep}, the existing facial attribute datasets are always limited because it is very hard to gather a completely and sufficiently labeled dataset. In Table~\ref{tab:data}, we list detailed information of various facial attribute datasets. For example, there are only 2476 training images in the ChaLearn15 apparent age estimation challenge~\cite{escalera2015chalearn}. Second, the number of images is very imbalanced in different label groups. What is more, distributions of different datasets are also very different. As Fig.~\ref{fig:datadis} depicts, there are two peaks on ChaLearn16~(the early one at around 2 years old and the latter one at 26 years old) and Morph~(the early one at around 20 years old and the latter one at 40 years old), while ChaLearn15  has only one peak. Similar phenomenon also appears in facial attractiveness datasets. These imbalances bring a serious challenge for developing an unbiased estimation system. Third, compared to other facial attributes, such as gender or expression, age/attractiveness estimation is a very fine-grained recognition task, \eg, we {humans} very hardly sense the change of one person's facial characteristics when he/she grew from 25 to 26 years old.

{Can we use a unified framework to effectively estimate age and attractiveness from facial images? 
First, these two tasks are both based on facial images thus it is possible to use a unified base model. Second, the labels are ordinal in either the age estimation or the facial attractiveness one. Generally, they are both integrated into ordinal regression or ordinal classification topics whose main character is that the categories are related in a natural or implied order. Common examples of such tasks are movie and facial attractiveness ratings (\eg, from 1 star to 5 stars) or facial age (\eg, from baby to child to adults) or image aesthetics~(\eg, from ``unacceptable" to  ``professional'' to ``exceptional"). Third, there is uncertainty information between any two adjacent labels for facial age and attractiveness tasks and their evaluation metric are also the same commonly using MAE. Therefore, we try to integrate facial age and attractiveness estimation into a unified framework in this paper.}

The common evaluation metric of age/attractiveness estimation is the Mean Absolute Error~(MAE) between the predicted value and ground-truth. Thus, it is very natural to treat facial attributes estimation as a metric regression problem~\cite{ranjan2017all} which minimizes the MAE. However, such methods usually cannot achieve satisfactory performance because some outliers may cause a large error term, which leads to an unstable training procedure. Later, Rothe~\etal~\cite{rothe2016deep} trained deep convolutional neural network~(CNN) for age estimation as multi-class classification, which maximizes the probability of ground-truth class without considering other classes. This method easily falls into over-fitting because of the imbalance problem among classes and limited training images~\cite{gao2017deep}.

Recently, ranking CNN~\cite{niu2016ordinal,chen2017using,Chen2017Deep, li2017d2c} and deep label distribution learning~(DLDL)~\cite{gao2017deep,fan2017label} techniques achieved state-of-the-art performance on facial age estimation. Both methods use the correlation information among adjacent ages at different levels. The ranking method transforms single-value estimation to a series of binary classification problems at the training stage. Then, the output of the rankers are aggregated directly from these binary outputs at predication stage. DLDL firstly converts a real-value to a discrete label distribution. Then, the aim of the training is to fit the entire distribution. At inference stage, like~\cite{rothe2016deep}, an expected value over the predicted distribution is taken as the final output. We can easily find that there is an inconsistency between the training objectives and the evaluation metric in all these methods. Thus, they may be suboptimal. We expect to improve their performance if this inconsistency is removed.

In addition, we observe that almost all state-of-the-art facial attributes estimation methods~\cite{rodriguez2017age,fan2017label,rothe2016deep,gao2017deep,antipov2016apparent} are initialized by a pre-trained model which is trained on large-scale image classification~(\eg, ImageNet~\cite{deng2009imagenet}) or face recognition~(\eg, VGGFace~\cite{parkhi2015deep}) datasets, and fine-tuned on the target dataset. These pre-trained models adopt some popular and powerful architectures~(\eg, VGGNet~\cite{simonyan2015very}). Unfortunately, these models often have huge computational cost and storage overhead. Taking VGG-16 for example, it has 138.34 million parameters, taking up more than 500MB storage space. Therefore, it is hard to be deployed on resource-constrained devices, \eg, mobile phones. Recently, some researchers devoted to compressing these pre-trained models so that both reducing the number of parameters and keeping accuracy are possible~\cite{iccv2017ThiNet}. Unlike these compression methods, we directly design a thin and deep network architecture and train it from scratch.

\begin{table}
 \centering
 \small
 \caption{Details of facial age and attractiveness datasets.
 }\label{tab:data}
 \scalebox{1.0}{
 \begin{tabular}{|@{\;}c@{\;}| @{\;}l@{\;}| @{\;}c@{\;}| @{\;}c@{\;}|}
  \hline
    &Dataset &\#Images &Label range\\
  \hline
  \multirow{4}*{Age}  
  &ChaLearn15~\cite{escalera2015chalearn}  &2476+1136  &3-85\\
  &ChaLearn16~\cite{escalera2016chalearn}  &5613+1978  &0-96\\
  &UTKFace~\cite{zhang2017age}      &24108  &0-116\\
  &Morph~\cite{ricanek2006morph}             &55134    &16-70\\
  \hline
  \multirow{2}*{Attractiveness} 
   &SCUT-FBP~\cite{xie2015scut}   &500 &1-5 \\
   &CFD~\cite{ma2015chicago}      &597 &1-7 \\
  \hline
 \end{tabular}}
\end{table}

In this paper, we integrate label distribution learning~\cite{geng2016label} and expectation regression into a unified framework to alleviate the inconsistency between training and evaluation stages with a simple and lightweight CNN architecture. The proposed approach effectively and efficiently improves the performance of the previous DLDL on both prediction error and inference speed for facial attributes estimation, so we call it DLDL-v2. Our contributions are summarized as follows.

\squishlist
 \item We provide, to the best of our knowledge, the first analysis and show that the ranking method is in fact learning label distribution implicitly. This result thus unifies existing state-of-the-art facial attributes estimation methods into the DLDL framework;
 \item We propose an end-to-end learning framework which jointly learns label distribution with the correlation information among neighboring labels and regresses single label ground-truth in both feature learning and classifier learning;
 \item We create new state-of-the-art results on facial age and attractiveness estimation tasks using single and small model without external age/attractiveness labeled data or multi-model ensemble;
 \item Our proposed framework is partly interpretable. We find the network employ different patterns to estimate age for people at different age stage. Meanwhile, we also quantitatively analyze the sensitivity of our approach to different face regions.
\squishend

We organize the rest of this paper as follows. Related works on facial attributes~(\eg, age and attractiveness) estimation are introduced in Section~\ref{sec:rw}. Then, Section~\ref{sec:oa} presents the proposed DLDL-v2 approach including the problem definition, the relationship between existing methods, and our joint learning framework and its model architecture. After that, the experiments are reported in Section~\ref{sec:ex}. In Section~\ref{sec:ud}, we discuss how DLDL-v2 makes the final determination for an input facial image and analyze why it can work well. Finally, the conclusion is given in Section~\ref{sec:co}. Some preliminary results have been published in a conference presentation~\cite{gao2018dldlv2}.

\section{Related Works}\label{sec:rw}
In the past two decades, many researchers have worked on facial attributes estimation. Earlier researches are two stage solutions, including feature extraction and model learning. Recently, deep learning methods are proposed, which integrate both stages into an end-to-end framework. In this section, we briefly review these two types of frameworks.

\textbf{Two stage methods.}
The task of the first stage is how to extract discriminative features from facial images. Active appearance model (AAM)~\cite{cootes2001active} is the earliest method through extracting shape and appearance features of face images. Later, the Bio-inspired feature (BIF)~\cite{guo2009human}, as the most successful age feature, is widely used in age estimation. But, in face attractiveness analysis, geometric features~\cite{zhang2011quantitative} and texture features~\cite{kagian2007humanlike} depended on facial landmark positions are widely used, since the BIF feature may be suboptimal for facial attractiveness prediction. Obviously, the drawback of hand-designed features is that one needs to re-design a feature extraction method when facing a new task, which usually requires domain knowledge and a lot of efforts. The second stage is how to exactly estimate facial attributes using these designed features. Classification and regression models {are often} used to estimate facial attributes. The former includes k-nearest neighbors (KNN), multilayer perceptron (MLP) and support vector machine (SVM), and the latter contains quadratic regression, support vector regression (SVR) and soft-margin mixture regression~\cite{huang2017soft}. Instead of classification and regression, ranking techniques~\cite{chang2011ordinal,chen2013cumulative,Wang2015Relative,Li2015Human,Wan2018Auxiliary} utilize the ordinal information of age to learn a model for facial age estimation.

In addition, Geng~\etal ~proposed a label distribution learning~(LDL) approach to utilize the correlation among adjacent labels, which improved performance on age estimation~\cite{geng2013facial} and beauty sensing~\cite{rensense}. Recently, some improvements of LDL~\cite{xing2016logistic,he2017data} have been proposed. Xing~\etal~\cite{xing2016logistic} used logistic boosting regression instead of the maximum entropy model in LDL. Meanwhile, He~\etal~\cite{he2017data} generated age label distributions through weighted linear combination of the label of input image and that of its context-neighboring images. These methods only learn a classifier, but not the visual representations.

\textbf{Single stage methods.} Deep CNNs have achieved impressive performance on various visual recognition tasks. The greatest success is learning feature representations instead of using hand-crafted features via the single stage learning strategy. Existing facial attribute estimation techniques fall into four categories: metric regression~(MR)~\cite{ranjan2017all}, multi-class classification~(DEX)~\cite{rothe2016deep}, Ranking~\cite{niu2016ordinal,chen2017using,Chen2017Deep} and DLDL~\cite{gao2017deep}. 

MR treats age estimation as a real-valued regression problem. The training procedure usually minimizes the squared difference between the estimated value and the ground-truth.

DEX adopts a general image classification framework which maximizes the probability of the ground-truth class during training. In the inference stage, Rothe~\etal~\cite{rothe2016deep} empirically showed that the expected value over the softmax-normalized output probabilities can achieve better performance than the class prediction of maximum probabilities. However, both MR and DEX easily lead to an unstable training~\cite{gao2017deep}.

Ranking methods transform facial attribute regression as a series of binary classification problems. Niu~\etal~\cite{niu2016ordinal} proposed a multi-out CNN via integrating multiple binary classification problems to a CNN. Then, Chen~\etal ~\cite{chen2017using,Chen2017Deep} trained a series of binary classification CNNs to get better performance. Given a testing image, the output of the rankers are aggregated directly from these binary outputs. 

DLDL converts a single value to a label distribution and learns it in an end-to-end fashion. Recently, Shen~\etal~\cite{shen2017label} proposed LDLFs via combining DLDL and differentiable decision trees. Hu~\etal~\cite{hu2017facial} exploited age difference information to improve the age estimation accuracy. These approaches have achieved state-of-the-art performance on age estimation. In addition, Yang~\etal~\cite{yangjoint} proposed a multi-task deep framework via jointly optimizing image classification and distribution learning for emotion recognition. However, these methods may be suboptimal, because there is an inconsistency between the training objectives and evaluation metric.

In this paper, we focus on how to alleviate or remove this inconsistency in a deep CNN with fewer parameters. Age and attractiveness estimation from still face images are suitable applications of the proposed research.

\section{Our Approach}\label{sec:oa}
In this section, we firstly give the definition of the joint learning  problem. Next, we show that ranking is implicitly learning label distribution. Finally, we present our framework and network architecture.

\subsection{The Joint Learning Problem}
\textbf{Notation.} We use boldface lowercase letters like $\vec p$ to denote vectors, and the $i$-th element of $\vec p$ is denoted as $p_i$. $\vec 1$ denotes a vector of ones. Boldface uppercase letters like $\mathbf{W}$ are used to denote matrices, and the element in the $i$-th row and $j$-th column is denoted as $W_{ij}$. The circle operator $\circ$ is used to denote element-wise multiplication.

The input space is $\mathcal{X}= \mathcal{R}^{h\times w \times c}$, where $h$, $w$ and $c$ are height, width and the number of channels of an input image, respectively. Label space $\mathcal{Y}= \mathcal{R}$ is real-valued. A training set with $N$ instances is denoted as $D=\{(\mathrm x^n, \mathrm y^n)\}_{n=1}^{N}$, where $\mathrm x^n\in \mathcal{X}$ denotes the $n$-th input image and $\mathrm y^n\in \mathcal{Y}$ its corresponding label. We may omit the image index $n$ for clarity. The joint learning aims to learn a mapping function $\mathcal F:\mathcal{X}\rightarrow\mathcal{Y}$ such that the error between prediction $\hat {\mathrm y}$ and ground-truth $\mathrm y$ be as small as possible on a given input image $\mathrm x$. 

However, metric regression often cannot achieve satisfactory performance. We observe that people usually predict another person's apparent age in a way like ``around 25 years old'' in real life, which indicates using not only 25 but also neighboring ages~(\eg, 24 and 26) to describe the face. Similar case also happens in facial attractiveness assessment. Based on the observation, label distribution learning methods can utilize the information via transforming the single value regression problem to a label distribution learning problem. 

To fulfill this goal, instead of outputting a single value $\mathrm y \in \mathcal R$  for an input $\mathrm x$, we quantize the range of possible $\mathrm y$ values into several labels. For example, it is reasonable to assume that $\mathrm y \in [0,100]$ in age estimation. Thus, we can define $\vec l= [0:\bigtriangleup l:100]$~(MATLAB notation) as the ordered label vector, where $\bigtriangleup l$ is a fixed real number. A label distribution $\vec p$ is then $(p_1, p_2, \ldots, p_K)$, where $p_i$ is the probability that $y = l_i$ (i.e., $\Pr(y = l_i )$ for $1 \leq i \leq K$)~\cite{gao2017deep}. Since we use equal step size $\bigtriangleup l$ in quantizing $y$, the probability density function~(p.d.f.) of normal distribution is a natural choice to generate the ground-truth $\vec p$ from $y$ and $\sigma$:
\begin{equation}
 p_k = \frac{1}{\sqrt{2\pi}\sigma} \exp\left( -\frac{(l_k-y)^2}{2 \sigma^2} \right) \,,  \label{eq:npdf}
\end{equation}
where $\sigma$ is a hyper-parameter. The goal of label distribution learning is to maximize the similarity between $\vec p$ and the CNN generated distribution $\vec {\hat p}$ at training stage. In the prediction stage, predicted distribution $\vec {\hat p}$ is reversed to a single value by a special inference function. It is suboptimal because there exists inconsistence between training objective and evaluation metric. We are  interested to not only learn the label distribution $\vec p$ but also regress a real value $y$ in one framework in an end-to-end manner.

\subsection{Ranking is Learning Label Distribution}\label{subsec:rem}
The ranking-based~\cite{niu2016ordinal,chen2017using,Chen2017Deep} and DLDL-based~\cite{gao2017deep,shen2017deep,shen2017label,fan2017label} methods have achieved state-of-the-art performance in facial age/attractiveness estimation problems. In this section, we analyze the essential relationship between them. 

\begin{figure}[t]
 \centering
 \subfloat[p.d.f.]
 {\includegraphics[height= 0.25\columnwidth]{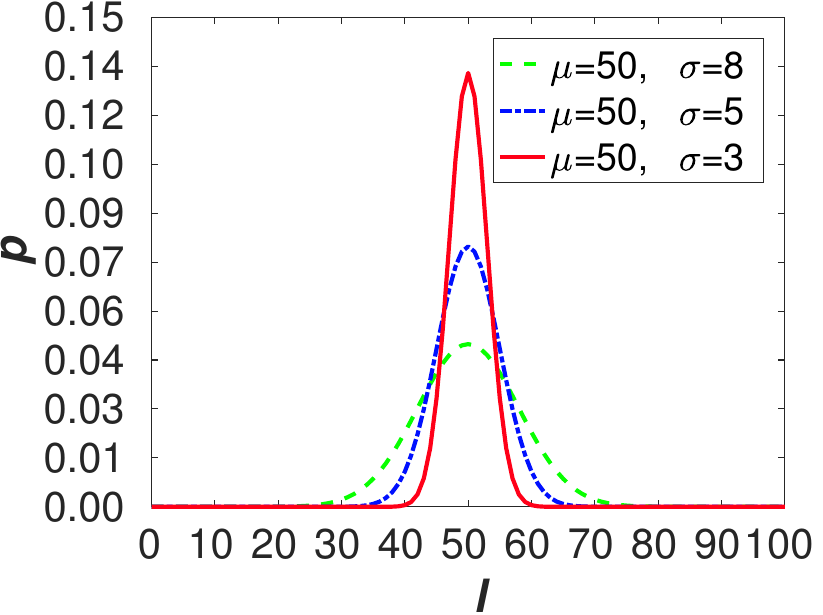}\label{fig:pdf}}
 \subfloat[c.d.f.]
 {\includegraphics[height= 0.25\columnwidth]{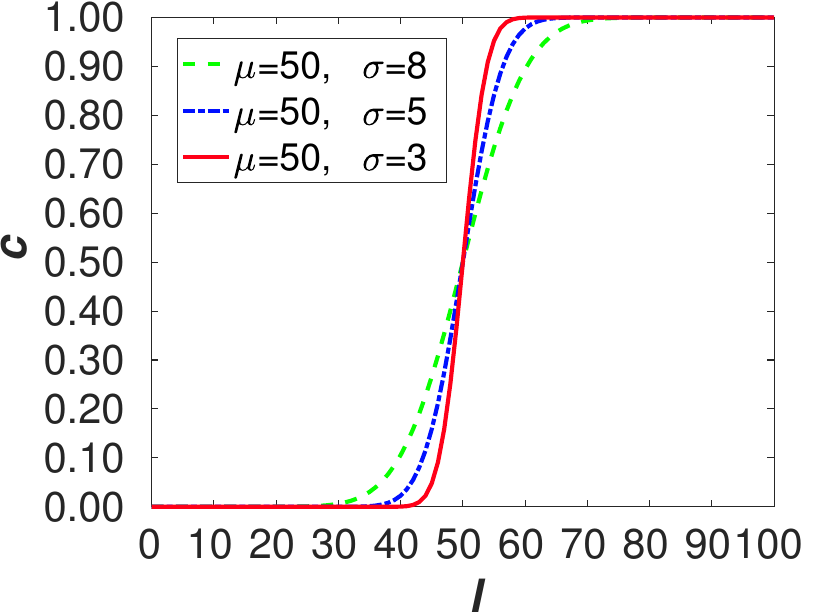}\label{fig:cdf}}
 \subfloat[one minus c.d.f.]
 {\includegraphics[height= 0.25\columnwidth]{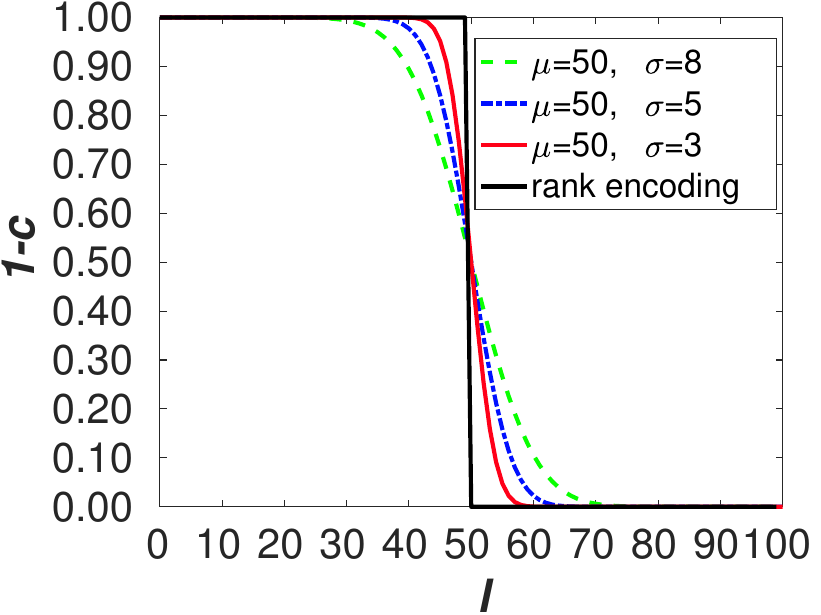}\label{fig:1-cdf-rank}}
\caption{(a) and (b) show p.d.f. and c.d.f. curves with the same mean and different standard deviation. (c) shows the curves of one minus c.d.f. and ranking encoding~(Best viewed in color).} \label{fig:pdf-cdf}
\vspace{-10pt}
\end{figure}

We explore their relationship from the perspective of label encoding. In DLDL-based approaches, for a face image $\mathrm x$ with true label $\mathrm y$ and hyper-parameter $\sigma$, the target vector $\vec p^{ld}$~(\ie, label distribution) is generated by a normal p.d.f.~(Eq.~\eqref{eq:npdf}). For example, the target vector of a 50 years old face is shown in Fig.~\ref{fig:pdf}, where $\vec l =[0,1,\ldots, 100]$. In ranking CNN, $K-1$ binary classifiers are required for $K$ ranks because the $k$-th binary classifier focuses on determining whether the age rank of an image is greater than $l_k$ or not. For a face image $\mathrm x$ with true label $\mathrm y \in (l_{k-1}, l_k]$, the target vector with length $K-1$ is encoded as $\vec p^{rank} =[1,\ldots,1,0,\ldots, 0]$, where the first $k-1$ values are 1 and the rest being 0. The target ranking vector of a 50 years old face is shown in Fig.~\ref{fig:1-cdf-rank} as the dark line. 

As we all know, for a generic normal distribution with p.d.f. $\vec p$, mean $\mathrm y$ and deviation $\sigma$, the cumulative distribution function~(c.d.f.) is
\begin{equation}
 c_k = \frac{1}{2}\Bigg[1 + \erf\bigg(\frac{l_k-\mathrm y}{\sigma\sqrt 2}\bigg)\Bigg]\,,  \label{eq:ncdf}
\end{equation}
where $\erf(x) = \frac{2}{\sqrt \pi} \int_0^x \mathrm{e}^{-t^2}\,\mathrm{d}t$. 
Fig.~\ref{fig:cdf} shows the c.d.f. corresponding to the p.d.f. in Fig.~\ref{fig:pdf}. From Eq.~\eqref{eq:ncdf}, we know
\begin{equation}
\begin{cases}
 1-c_k > 0.5, & \quad \text{if } l_k<\mathrm y \\
 1-c_k \leq 0.5, & \quad \text{if } l_k\geq\mathrm y 
 \end{cases}\,,
\end{equation}
As shown in Fig.~\ref{fig:1-cdf-rank}, the curve of $\vec 1-\vec c$ is very close to that of $\vec p^{rank}$ when $\sigma$ is a small positive real number. Thus, \begin{equation}\label{eq:rs}
 p_k^{rank} \approx 1 - c_k \,,
\end{equation}
where $k= 1,2,\ldots,K-1$. 

\emph{Eq.~\eqref{eq:rs} shows $\vec p^{rank}$ is a specific case of label distribution learning, where the distribution is the cumulative one with $\sigma \to 0$}. That is to say, Ranking is to learn a c.d.f. essentially, while DLDL aims at learning a p.d.f. More generally, we have
\begin{equation}
\vec c = \mathbf{T}\vec p^{ld} \label{eq:pdftocdf} \,,
\end{equation}
where $\mathbf{T}$ is a transformation matrix with  $T_{ij}=1$ for all $i\le j$ and $T_{ij}=0$ when $i>j$. Substituting~\eqref{eq:pdftocdf} in to~\eqref{eq:rs}, we have
\begin{equation}
\vec p^{rank}  \approx \vec 1 - \mathbf{T}\vec p^{ld} \label{eq:cdftopdf} \,.
\end{equation}
Therefore, there is a linear relationship between Ranking encoding and label distribution. The label distribution encoding $\vec p^{ld}$ can represent more meaningful age/attractiveness information with different $\sigma$, but ranking encoding  $\vec p^{rank}$ does not. Furthermore, DLDL is more efficient, because only one network has to be trained. 

However, as discussed earlier, all these methods may be suboptimal because there exists inconsistency between training objective and evaluation metric.

\begin{figure}
 \centering
 \includegraphics[width= 1\columnwidth]{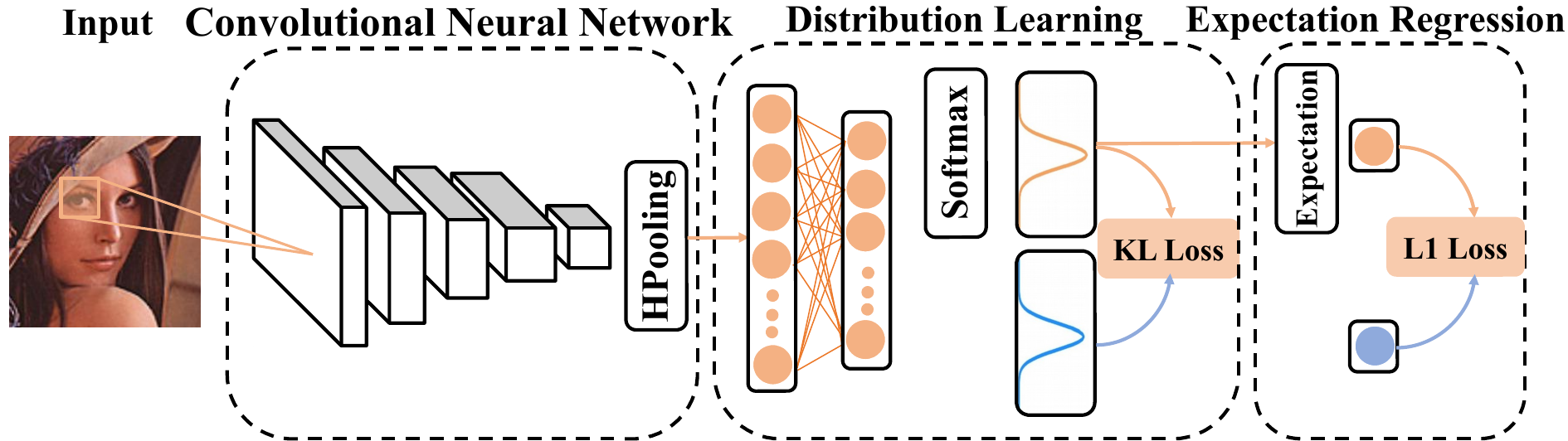}
\caption{The framework of our DLDL-v2. Given an input image and its ground-truth, we firstly generate a label distribution and then jointly optimize label distribution learning and expectation regression in one unified framework in an end-to-end manner.} \label{fig:dldlv2f}
\vspace{-5pt}
\end{figure}

\subsection{Joint Learning Framework}
In order to jointly learn label distribution and output the expectation, in this section we propose the DLDL-v2 framework.

\subsubsection{The Label Distribution Learning Module}
In order to utilize the good properties of label distribution learning, we integrate it into our framework to formulate a label distribution learning module. As shown in Fig.~\ref{fig:dldlv2f}, this module includes a fully connected layer, a softmax layer and a loss layer. This module follows the DLDL method in~\cite{gao2017deep}.

Specifically, given an input image $\mathrm x$ and the corresponding label distribution $\vec p$, we assume $\vec f = \mathcal{F}(\mathrm x;\vec \theta)$ is the activation of the last layer of CNN, where $\vec \theta$ denotes the parameters of the CNN. A fully connected layer transfers 
$\vec f$ to $\vec x\in \mathcal {R}^K$ by
\begin{equation}\label{eq:linear}
\vec x = \mathbf{W}^{T}\vec f + \vec b\,.
\end{equation}
Then, we use a softmax function to turn $\vec x$ into a probability distribution, that is,
\begin{equation}\label{eq:sm}
\hat p_k = \frac{\exp (x_k)}{\sum_t {\exp (x_t)}}\,.
\end{equation}
Given an input image, the goal of the label distribution learning module is to find $\vec \theta$, $\mathbf W$, and $\vec b$ to generate $\hat {\vec p}$ that is similar to $\vec p$.

We employ the Kullback-Leibler divergence as the measurement of the dissimilarity between ground-truth label distribution and prediction distribution. Thus, we can define a loss function on one training sample as follows~\cite{gao2017deep}:
\begin{equation}\label{eq:lld}
L_{ld} = \sum_k {p_k \ln \frac{p_k}{\hat {p_k}}}\,.
\end{equation}

\subsubsection{The Expectation Regression Module}
Note that the label distribution learning module only learns a label distribution but cannot regress a precise value. In order to reduce the inconsistency between training and evaluation stages, we propose an expectation regression module to further refine the predicted value. As shown in Fig.~\ref{fig:dldlv2f}, this module includes an expectation layer and a loss layer.

The expectation layer takes the predicted distribution and label set as input and emits its expectation 
\begin{equation}\label{eq:exp}
\hat y = \sum_k \hat p_k l_k \,,
\end{equation}
where $\hat p_k$ denotes the prediction probability that the input image belongs to label $l_k$.
Given an input image, the expectation regression module minimizes the error between the expected value $\hat y$ and ground-truth $y$. We use the $\ell_1$ loss as the error measurement as follows:
\begin{equation}\label{eq:lep}
L_{er} = |\hat y - y|\,,
\end{equation}
where $|\cdot|$ denotes absolute value. Note that this module does not introduce any new parameter.

\subsubsection{Learning}
Given a training data set $D$, the learning goal of our framework is to find $\vec \theta$, $\mathbf W$ and $\vec b$ via jointly learning label distribution and expectation regression. Thus, our final loss function is a weighted combination of the label distribution loss $L_{ld}$ and the expectation regression loss $L_{er}$:
\begin{equation}\label{eq:loss1}
L = L_{ld} + \lambda L_{er}\,,
\end{equation}
where $\lambda$ is a weight which balances the importance between two types of losses.
Substituting Eq.~\eqref{eq:lld}, Eq.~\eqref{eq:exp} and Eq.~\eqref{eq:lep} into~Eq.~\eqref{eq:loss1}, we have
\begin{equation}\label{eq:loss2}
L = -\sum_k p_k\ln {\hat p_k} + \lambda \Big|\sum_k{\hat {p_k}l_k}-y\Big| \,.
\end{equation}

We adopt stochastic gradient descent to optimize parameters of our model. The derivative of $L$ with respect to $\hat p_k$ is
\begin{equation}\label{eq:loss}
\frac {\partial {L}}{\partial{\hat p_k}} = -\frac{p_k}{\hat p_k} + \lambda l_k \sign(\hat y- y)\,.
\end{equation}
For any $k$ and $j$, the derivative of softmax (Eq.~\eqref{eq:sm}) is well known, as
\begin{equation}
\frac {\partial {\hat p_k}}{\partial{x_j}} = \hat p_k(\delta_{(k=j)}-\hat p_j)\,,
\end{equation}
where $\delta_{(k=j)}$ is 1 if $k=j$, and 0 otherwise. According to the chain rule, we have
\begin{equation}
\frac {\partial L}{\partial{x_j}} = (\hat p_j -p_j) + \lambda \text{sign}(\hat y- y)\hat p_j(l_j-\hat y)\,,
\end{equation}
\begin{equation}
\text{\emph{i.e.},}~~\frac {\partial L}{\partial \vec x} = \hat{\vec p} -\vec p + \lambda \sign(\hat y- y)\vec {\hat p}\circ(\vec l-\hat y\vec 1)\,.
\end{equation}
Applying the chain rule for Eq.~\eqref{eq:linear} again, the derivative of $L$ with respect to $\mathbf W$, $\vec b$ and $\vec \theta$ are easily obtained, as
\begin{equation}
\frac {\partial L}{\partial{\mathbf W}} =\frac {\partial L}{\partial \vec x}\vec f,\\
\frac {\partial L}{\partial{\vec b}} =\frac {\partial L}{\partial \vec x},\\
\frac {\partial L}{\partial{\vec 
\theta}}  =\frac {\partial L}{\partial \vec x}\mathbf W^{T}\frac{\partial \mathcal{F}}{\partial \vec \theta}\,.
\end{equation}

Once $\mathbf W$, $\vec b$ and $\vec \theta$ are learned, the prediction value $\hat y$ of any new
instance $\mathrm x$ is generated by Eq.~\eqref{eq:exp} in a forward network computation.

\subsection{Network Architecture}
Considering both model size and efficiency, we modify VGG-16~\cite{simonyan2015very} from four aspects as follows. VGG-16 consists of 13 convolution~(Conv) layers, five max-pooling~(MP) layers and three fully connected~(FC) layers, and each Conv layer and FC layer is followed by a ReLU layer.

First, we observe that the three FC layers roughly contain $90\%$ parameters of the whole model. We remove all FC layers and add a hybrid-pooling~(HP) layer which is constructed by an MP layer and a global avg-pooling~(GAP) layer. We find that the HP strategy is more effective than single GAP. Second, to further reduce model size, we reduce the number of the filters in each Conv layer to make it thinner. Third, batch normalization~(BN)~\cite{ioffe2015batch} has been widely used in the latest architecture such as ResNet~\cite{he2016deep}. Thus, we add a BN layer after each Conv layer to accelerate network training. Last but not least, we add the label distribution learning module and the expectation regression module after the HP layer, as shown in Fig.~\ref{fig:dldlv2f}.

Since we design the network for age/attractiveness estimation and its architecture is thinner than the original VGG-16, we call our model ThinAgeNet or ThinAttNet which employs the compression rate of 0.5 and has 3.7M parameters. \footnote {0.5 compression rate means every Conv layer has only 50\% channels as that in VGG-16.} 
We also train a very small model with the compression rate of 0.25, and we call it TinyAgeNet or TinyAttNet which only has 0.9M parameters.

\section{Experiments}\label{sec:ex}
In this section, we conduct experiments to validate the effectiveness of the proposed DLDL-v2 approach on facial age and attractiveness datasets, based on the open source framework Torch7. All experiments are conducted on an NVIDIA M40 GPU. In order to re-produce all results in this paper, we will release source code and pre-trained models upon paper acceptance.

\subsection{Implementation Details}

\textbf{Pre-preprocessing.} We use multi-task cascaded CNN~\cite{zhang2016joint} to conduct face detection and facial points detection for all images. Then, based on these facial points, we align faces to the upright pose. Finally, all faces are cropped and resized to $224\times224$. Before feeding to the network, all resized images are to subtract mean and divide standard deviation for each color channel.

\textbf{Data Augmentation.} There are many non-controlled environmental factors such as face position, illumination, diverse backgrounds, image color~(\ie, gray and color) and image quality, especially in the ChaLearn datasets. To handle these issues, we apply data augmentation techniques to every training image, so that the network can take a different variation of the original image as input at each epoch of training. Specifically, we mainly employ five types of augmentation methods for a cropped and resized training image, including random horizontal flipping, random scaling, random color/gray changing, random rotation and standard color jittering.

\textbf{Training Details.} We pre-train a deep CNN model with softmax loss for face recognition on a subset of the MS-Celeb-1M dataset~\cite{guo2016ms}. One issue is that a small part of identities have a large number of images and others have only a few in this dataset. To avoid the imbalance problem among identities, we cut those identities whose number of images is lower than a threshold. In our experiments, we use about 5M images of 54K identities as training data.

After pre-training is finished, we remove the classification layer of the network and add the label distribution learning and expectation regression modules. Then, fine-tuning is conducted on target datasets. We set $\lambda=1$ in Eq.~\eqref{eq:loss}. The ordered label vector is defined as $\vec l =[l_{min}:\bigtriangleup l:l_{max}]$ (MATLAB notation). For age estimation, we set $l_{min}=0$, $\bigtriangleup l=1$, and $l_{max}=100$. For attractiveness estimation, we set $l_{min}=1$ and $\bigtriangleup l=0.1$. Because there are different scoring rules on SCUT-FBP and CFD dataset, $l_{max}$ is set to 5 and 7, respectively. The label distribution of each image is generated using Eq.~\eqref{eq:npdf}. The ground-truth~(age or attractiveness score) is provided in all datasets. The standard deviation, however, is provided in ChaLearn15, ChaLearn16 and SCUT-FBP, but not Morph and CFD. {Generally, a standard deviation $\sigma$ that is close to the interval between neighboring labels is a good choice that has been discussed and analyzed in the DLDL~\cite{gao2017deep}.  Following this principle,} we simply set $\sigma=2$ in Morph and $\sigma=0.5$ in CFD, respectively. All networks are optimized by \emph{Adam}, with $\beta_1=0.9$, $\beta_2=0.999$ and $\epsilon=10^{-8}$. The initial learning rate is 0.001 for all models, and it is decreased by a factor of 10 every 30 epochs. Each model is trained 60 epochs using mini-batches of 128.

\textbf{Inference Details.} At the inference stage, we feed a testing image and its horizontally flipping copy into the network and average their predictions as the final estimation for the image.
\vspace{-8pt}

\subsection{Evaluation Metrics}
MAE is used to evaluate the performance of facial age or attractiveness estimation,
\begin{equation}
\text{MAE} = \frac{1}{N}\sum_{n=1}^{N}|\hat y^n - y^n|,
\end{equation}
where $\hat y^n$ and  $y^n$ are the estimated and the ground-truth of the $n$-th testing image, respectively. In addition, a special measurement ($\epsilon$-error) is defined by the ChaLearn competition, as
\begin{equation}
\epsilon\text{-error} = \frac{1}{N}\sum_{n=1}^N \Bigg[1 - \exp\left(-\frac{(\hat{y}^n- y^n)^2}{2(\sigma^n)^2}\right)\Bigg],   \label{eq-me}
\end{equation}
where $\sigma^n$ is the standard deviation of the $n$-th testing image.

We also follow~\cite{xie2015scut,fan2017label} to compute Root Mean Squared Error (RMSE) and Pearson Correlation (PC), which can be computed as:
\begin{equation}
\text{RMSE} = \sqrt{\frac{1}{N}\sum_{n=1}^{N}{|\hat y^n - y^n|}^2},
\end{equation}
\begin{equation}
\text{PC} = \frac{\sum_{n=1}^{N}(y^n-\overline y)(\hat y^n-\overline{\hat y})}{\sqrt{\sum_{n=1}^{N}{|y^n - \overline y|}^2}\sqrt{\sum_{n=1}^{N}{|\hat y^n - \overline{\hat y}|}^2}},
\end{equation}
where $\overline y = \frac{1}{N}\sum_{n=1}^{N}{y^n}$, and $\overline {\hat y} = \frac{1}{N}\sum_{n=1}^{N}{{\hat y}^n}$ are the mean values of the ground-truth and predicted scores over all testing images. These two evaluation metrics are only utilized to evaluate the performance of facial attractiveness estimation. 

\subsection{Experiments on Age Estimation}\label{sec:eage}
\subsubsection{Age Estimation Datasets}
{Three} types of datasets are used in our experiments. The first type contains two small-scale apparent age datasets~(ChaLearn15~\cite{escalera2015chalearn} and ChaLearn16~\cite{escalera2016chalearn}) which are collected in the wild. The second type is a large-scale real age dataset~(Morph)~\cite{ricanek2006morph}. We follow the experimental setting in~\cite{gao2018dldlv2} for evaluation. {The third one is UTKFace~\cite{zhang2017age} which the ground truth of age are estimated through the DEX algorithm and double checked by a human annotator.} We utilize the split employed in~\cite{cao2020rank}, with 20\% test images and 80\% images for training.

\subsubsection{Age Estimation Results}
We compare our approach with the state-of-the-art in both prediction performance and inference time. 

\textbf{Low Error.}
Table2~\ref{tab:sotamorph},~\ref{tab:sotachalearn15},~\ref{tab:sotachalearn16} and~
\ref{tab:sotautkface} report the comparisons of the MAE and $\epsilon$-error performance of our method and previous state-of-the-art methods on four age estimation datasets. 

In the ChaLearn15 challenge, the best result came from DEX. DEX method's success relies on a lot of external age labeled training images (260282 additional photos). Under the same setting~(without external data), our method outperforms DEX by a large margin in Table~\ref{tab:csage}.
On ChaLearn16, the $\epsilon$-error $0.267$ of our approach is closest to the best competition result 0.241~\cite{antipov2016apparent} on the testing set. Note that our result is only based on a single model without external age labeled data. In~\cite{antipov2016apparent}, they not only used external age labeled data but also employed multi-model ensemble. On Morph, our method creates a new state-of-the-art 1.969 MAE. To our best knowledge, this is the first time to achieve below two years in MAE on the Morph dataset. 

In short, our DLDL-v2~(ThinAgeNet) outperforms the state-of-the art methods without external age labeled data and multi-model ensemble on ChaLearn15, ChaLearn16, Morph and {UTKFace}.

\begin{table}[t]
 \centering
 \small
 \caption{Comparisons with state-of-the-art methods for real age estimation on Morph dataset.}\label{tab:sotamorph}
   \scalebox{0.80}{
 \begin{tabular}{|@{\;}l@{\;}|@{\;}r@{\;}|@{\;}c@{\;}|@{}c@{\;} |@{\;}c@{\;}|@{}c@{\;}|}
   \hline
  \multirow{2}*{Methods} &\multirow{2}*{Venue Year} &External  &\multicolumn{2}{c|}{Model}   &\multirow{2}{*}{MAE$\downarrow$ }
    \\\cline{4-5}  & &Data &~Single?&Network    & \\
 \hline
Human~\cite{han2015demographic} &TPAMI 2015  &$\times$     &Yes   &human workers     &6.30\\
OR-CNN~\cite{niu2016ordinal} &CVPR 2016  &$\times$    &Yes   &MR-CNN     &3.27\\
DEX \cite{rothe2016deep}~&IJCV 2016    &$\times$    &Yes   &VGG-16     &3.25\\
DEX \cite{rothe2016deep}~&IJCV 2016    &$\checkmark$    &Yes   &VGG-16     &2.68\\
LDAE~\cite{antipov2017effective} &PR 2017    &$\checkmark$     &Yes   &VGG-16     &2.35\\
DLDL \cite{gao2017deep}~&TIP 2017    &$\times$    &Yes  &VGG-Face     &2.42\\
Rank-CNN~\cite{chen2017using,Chen2017Deep} &CVPR 2017    &$\times$   &No  &Binary CNNs     &2.96\\
DLDLF~\cite{shen2017label}    &NIPS 2017    &$\times$    &Yes   &VGG-Face     &2.24\\
DRFS~\cite{shen2017deep}  &CVPR 2018    &$\times$    &Yes   &VGG-16     &2.17\\
MV~\cite{pan2018mean}  &CVPR 2018    & $\checkmark$   & Yes  &VGG-16    &2.16\\
AgeEn~\cite{tan2018efficient} &TPAMI 2018    &$\checkmark$    &Yes   &VGG-16    &2.52\\
C3AE~\cite{zhang2019c3ae} &CVPR 2019    &$\checkmark$    &Yes  &C3AE     &2.75\\
BridgeNet~\cite{li2019bridgenet}  &CVPR 2019    & $\checkmark$   & Yes  &VGG-16    &2.38\\
SADAL~\cite{liu2020similarity}&TMM 2020    &$\times$    &Yes   &VGG-Face    &2.59\\
VDAL~\cite{liu2020similarity} &TMM 2020    &$\times$   &Yes   &VGG-Face    &2.57\\
1CH~\cite{lim2020order}~&ICLR 2020    &$\checkmark$    &Yes   &VGG-16 (w/o FCs)   &2.22\\
POE~\cite{li2021learning} &CVPR 2021   &$\checkmark$   &Yes   &VGG-16     &2.35\\
PML~\cite{deng2021pml} &CVPR 2021   &$\times$   &Yes   &ResNet-34     &2.31\\
PML~\cite{deng2021pml} &CVPR 2021   &$\checkmark$    &Yes   &ResNet-34     &\underline{2.15}\\
\hline
 \multicolumn{2}{|c|} {DLDL-v2~(TinyAgeNet)}    &$\times$  &Yes   &Thin VGG-16   &2.29 \\
 \multicolumn{2}{|c|} {DLDL-v2~(ThinAgeNet)}    &$\times$ &Yes   &Tiny VGG-16    &\textBF{1.97}\\
  \hline
 \end{tabular}}
\end{table}

\begin{table}
\fontsize{9}{10}\selectfont
 \centering
 \small
 \caption{Comparisons with state-of-the-art methods for apparent age estimation on ChaLearn 2015 dataset.}\label{tab:sotachalearn15}
  \scalebox{0.84}{
 \begin{tabular}{|@{\;}l@{\;}|@{\;}r@{\;}|@{\;}c@{\;}|@{}c@{\;} |@{\;}c@{\;}|@{}c@{\;}@{}c@{\;}|}
  \hline
  \multirow{2}*{Methods} &\multirow{2}*{Venue Year} &External  &\multicolumn{2}{c|}{Model}   &\multicolumn{2}{c|}{ValSet}   
    \\\cline{4-7}  & &Data &~Single?&Network    & MAE$\downarrow$  &$\epsilon$-error$\downarrow$  \\
  \hline
Human~\cite{han2015demographic} &TPAMI 2015  &$\times$    &Yes   &ResNet-50     &- &0.340\\
DEX \cite{rothe2016deep}~&IJCV 2016    &$\times$    &Yes   &VGG-16     &5.369 &0.456\\
DEX \cite{rothe2016deep}~&IJCV 2016    &$\checkmark$   &Yes   &VGG-16     &3.252 &0.282\\
DLDL \cite{gao2017deep}~&TIP 2017    & $\times$   &Yes   &VGG-Face     &3.510 &0.310\\
AGEn~\cite{tan2018efficient} &TPAMI 2018    &$\checkmark$    &Yes   &VGG-16    &3.210 &0.280\\
ODL~\cite{liu2017ordinal} &TCSVT 2019    &$\times$    &Yes   &VGG-Face    &3.950 &0.312\\
SADAL~\cite{liu2020similarity} &TMM 2020    &$\times$   &Yes   &VGG-Face    &3.780 &0.309\\
VDAL~\cite{liu2020similarity}  &TMM 2020    &$\times$    &Yes   &VGG-Face      &3.580 &0.285\\
PML~\cite{deng2021pml}     &CVPR 2021   &$\times$      &Yes   &ResNet-34     &3.455 &0.293\\
PML~\cite{deng2021pml}     &CVPR 2021   &$\checkmark$    &Yes   &ResNet-34     &\textBF{2.915} &\textBF{0.243}\\
\hline
 \multicolumn{2}{|c|} {DLDL-v2~(TinyAgeNet)}    &$\times$  &Yes   &Tiny VGG-16   &3.427 &0.301 \\
 \multicolumn{2}{|c|} {DLDL-v2~(ThinAgeNet)}   &$\times$ &Yes   &Thin VGG-16     &\underline{3.135} &\underline{0.272}\\
\hline
\end{tabular}}
\end{table}
\begin{table}[t]
 \centering
 \small
 \caption{Comparisons with state-of-the-art methods for apparent age estimation on ChaLearn 2016 dataset.}\label{tab:sotachalearn16}
 \scalebox{0.81}{
 \begin{tabular}{|@{\;}l@{\;}|@{\;}r@{\;}|@{\;}c@{\;}|@{}c@{\;} |@{\;}c@{\;}|@{}c@{\;}@{}r@{\;}|}
  \hline
  \multirow{2}*{Methods} &\multirow{2}*{Venue Year}  &External  &\multicolumn{2}{c|}{Model}   &\multicolumn{2}{c|}{TestSet}
    \\\cline{4-7}  & &Data &Num. Nets$\downarrow$ &Network    & MAE$\downarrow$  &$\epsilon$-error$\downarrow$ \\
\hline
OrangeLabs~\cite{antipov2016apparent}   &CVPRW 2016   &$\checkmark$        &14  &VGG-16    &- &0.241 \\
{palm-seu}~\cite{huo2016deep}                 &CVPRW 2016   &$\checkmark$    &4   &VGG-Face   &- &0.321 \\
cmp+ETH~\cite{uricar2016structured}        &CVPRW 2016   &$\checkmark$    &10  &VGG-16   &- &0.336 \\
\hline
AGEn~\cite{tan2018efficient} &TPAMI 2018    &$\checkmark$    &1   &VGG-16    &3.820 &0.310\\
MV~\cite{pan2018mean}  &CVPR 2018    & $\checkmark$   & 1  &VGG-16    &- &0.287\\
\hline
  \multicolumn{2}{|c|} {DLDL-v2~(TinyAgeNet)}  &$\times$  &1  &Tiny-VGG   &3.765 &0.291 \\
 \multicolumn{2}{|c|} {DLDL-v2~(ThinAgeNet)}   &$\times$ &1   &Thin-VGG    &\textBF{3.452} &\textBF{0.267}\\
  \hline
 \end{tabular}}
 \end{table}
 
\begin{table}
 \centering
 \small
 \caption{Comparisons with state-of-the-art methods for age estimation on UTKFace. The results of OR-CNN~\cite{niu2016ordinal} and CORAL~\cite{cao2020rank} are from Gustafsson \etal~\cite{gustafsson2020energy}. \textsuperscript{\dag}indicates methods using entire UTKFace dataset (0-116 years old) and \textsuperscript{\ddag}indicates methods using a subset of UTKFace covering faces between 21 and 60 years old.}\label{tab:sotautkface}
  \scalebox{0.85}{
\begin{tabular}{|@{\;}l@{\;}| @{\;}r@{\;}| @{\;}c@{\;}|@{}c@{\;} | @{}c@{\;}  |*{1}{@{\;}c@{\;}}|}
  \hline
  \multirow{2}*{Methods} &\multirow{2}*{Venue Year} &External  &\multicolumn{2}{c|}{Model}    &\multirow{2}{*}{MAE$\downarrow$ }
 \\\cline{4-5}
   & &Data &~Single?&Network  & \\
  \hline
OR-CNN~\cite{niu2016ordinal}\textsuperscript{\ddag}   &CVPR 2016         &$\checkmark$    &Yes   &ResNet-50     &5.74\\
Andrey Savchenko \cite{savchenko2019efficient}\textsuperscript{\dag}   &PeerJ 2019            &$\checkmark$    &Yes   &MobileNet-v2     &5.44\\
Axel Berg \etal  \cite{berg2021deep}\textsuperscript{\ddag}  &ICPR 2019                           &$\times$  &Yes   &ResNet-50     &4.55\\
  CORAL~\cite{cao2020rank}\textsuperscript{\ddag}   &PRL 2020                 &$\times$  &Yes   &ResNet-50   &5.47 \\
 Gustafsson \etal~\cite{gustafsson2020energy}\textsuperscript{\ddag}   &ECCV 2020   &$\times$  &Yes   &ResNet-50     &4.65\\
    \hline
 \multicolumn{2}{|c|} {DLDL-v2~(TinyAgeNet)\textsuperscript{\ddag} }    &$\times$  &Yes   &Tiny-VGG   &\underline{4.26} \\
 \multicolumn{2}{|c|} {DLDL-v2~(ThinAgeNet)\textsuperscript{\ddag} }   &$\times$ &Yes   &Thin-VGG    &\textBF{4.24}\\
  \multicolumn{2}{|c|} {DLDL-v2~(TinyAgeNet)\textsuperscript{\dag}}     &$\times$  &Yes   &Tiny-VGG   &\underline{4.27} \\
 \multicolumn{2}{|c|} {DLDL-v2~(ThinAgeNet)\textsuperscript{\dag}}     &$\times$ &Yes   &Thin-VGG    &\textBF{4.15}\\
  \hline
 \end{tabular}}
\end{table}

\textbf{High Efficiency.} We measure the speed on one {P40 GPU accelerated by cuDNN v7.6}. 
The number of parameters, {forward/backward memory size, computational complexity (GFlops) and actual inference time}
of our approach and some previous methods are reported in Table~\ref{tab:sotanet}. Since~\cite{niu2016ordinal} and~\cite{chen2017using} do not release pre-trained models, we cannot test the running time and report the number of parameters of these models. \cite{gao2017deep,rothe2016deep,antipov2017effective,shen2017label,shen2017deep,pan2018mean,tan2018efficient,li2019bridgenet,liu2020similarity,li2021learning,liu2017ordinal,tan2018efficient,li2019bridgenet,liu2020similarity,li2021learning,liu2017ordinal} all used similar network architecture (\ie, VGG-16 or VGGFace). Since~\cite{antipov2016apparent} employed 14 models, it's model size and running time is 14 times of~\cite{rothe2016deep} and~\cite{gao2017deep}. Compared to the state-of-the-art, DLDL-v2~(ThinAgeNet) achieves the best performance using single model with 36$\times$ fewer parameters and 3$\times$ reduction in inference time. Furthermore, we also report DLDL-v2's TinyAgeNet results which achieve a better result~(150$\times$ fewer parameters and 6$\times$ speed improvement) than the original DLDL~\cite{gao2017deep}. 

\begin{table}[t]
 \centering
 \small
 \caption{Comparisons of model parameters and forward times with state-of-the-art methods. M means million ($10^6$), G means billion ($10^9$), MB means megabyte and Time denotes the average forward times of $n$ images in milliseconds on one P40 GPU.}\label{tab:sotanet}
    \scalebox{0.86}{
\begin{tabular}{|@{\;}l@{\;}|@{}c@{\;} | @{}r@{\;}  | @{}r@{\;}   | @{}r@{\;}   | @{}c@{\;} |*{1}{@{\;}c@{\;}}|}
  \hline
  \multirow{2}*{Methods}  &{Model}   &Param  &Memory  &GFlops   &\multicolumn{2}{c|}{Time (ms)$\downarrow$ } \\\cline{6-7}
   &artitecture   &~(M)$\downarrow$~ &~(MB)$\downarrow$~  &(G)$\downarrow$~  &~~n=1~ &~n=128\\
  \hline
  \cite{gao2017deep,rothe2016deep,antipov2017effective,shen2017label,shen2017deep,pan2018mean}         &\multirow{3}*{VGG-16}      &\multirow{3}*{134.7}  &\multirow{3}*{322.13}     &\multirow{3}*{15.53} &\multirow{3}*{~7.09} &\multirow{3}*{3.36}\\
  \cite{tan2018efficient,li2019bridgenet,liu2020similarity,li2021learning,liu2017ordinal}                &       & &    & &~ &\\
  \cite{antipov2016apparent,huo2016deep,uricar2016structured}                       &       & &    & &~ &\\
  \cite{lim2020order}           &VGG-16(w/o FCs)      &14.8    &321.76  &15.41  &~~4.93 &3.26\\
  \cite{gustafsson2020energy,berg2021deep,han2015demographic}            &ResNet-50                  &23.7  &286.55   &4.12  &~~8.15 &1.92\\
  \cite{deng2021pml}                          &ResNet-34                  &21.3   &96.28  &3.67 &~~6.00 &0.88\\
  \cite{savchenko2019efficient}                          &MobileNet-v2               &2.4     &152.86 &\textbf{0.32}  &~~6.48 &0.87\\
   \hline
 ThinAgeNet   &Thin-VGG                               & 3.7  &160.88  &3.88 &~~2.35 &1.17 \\
 TinyAgeNet   &Tiny-VGG                               &\textbf{0.9}  &\textbf{80.44}  &0.98 &\textbf{~~2.20}&\textbf{0.53}\\
  \hline
 \end{tabular}}
\end{table}

\begin{figure*}
	\captionsetup[subfigure]{labelformat=empty}
	\captionsetup[subfigure]{justification=centering}
	\centering
	\vspace{-10pt}
	\subfloat[]   {{Input}}\hspace{2pt}
	\subfloat[] {\includegraphics[width= 0.110\columnwidth,keepaspectratio]{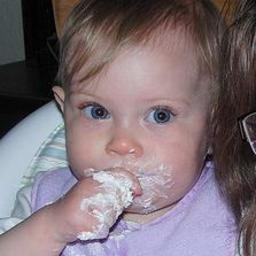}}
	\subfloat[] {\includegraphics[width= 0.110\columnwidth,keepaspectratio]{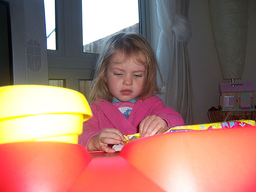}}
	\subfloat[] {\includegraphics[width= 0.110\columnwidth,keepaspectratio]{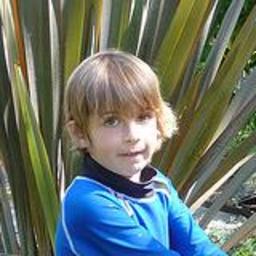}}
	\subfloat[] {\includegraphics[width= 0.110\columnwidth,keepaspectratio]{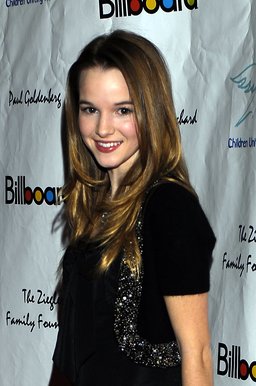}}
	\subfloat[] {\includegraphics[width= 0.110\columnwidth,keepaspectratio]{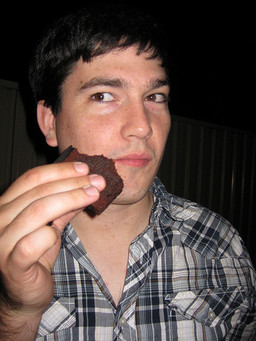}}
	\subfloat[] {\includegraphics[width= 0.110\columnwidth,keepaspectratio]{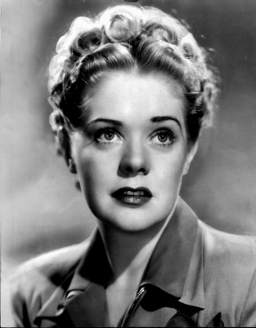}}
	\subfloat[] {\includegraphics[width= 0.110\columnwidth,keepaspectratio]{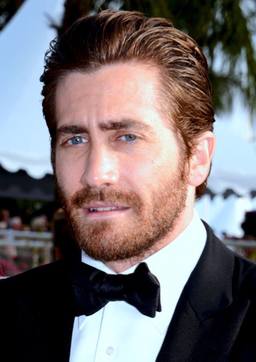}}
	\subfloat[] {\includegraphics[width= 0.110\columnwidth,keepaspectratio]{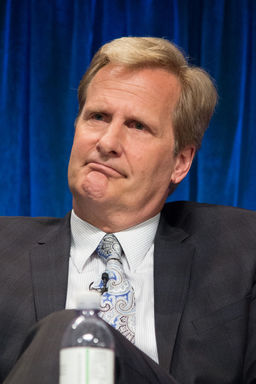}}
	\subfloat[] {\includegraphics[width= 0.110\columnwidth,keepaspectratio]{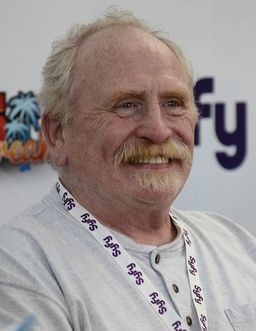}}
	\subfloat[] {\includegraphics[width= 0.110\columnwidth,keepaspectratio]{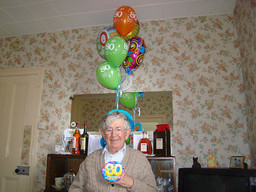}}
	\quad  \quad
	\subfloat[] {\includegraphics[width= 0.110\columnwidth,keepaspectratio]{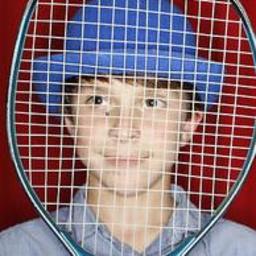}}
	\subfloat[] {\includegraphics[width= 0.110\columnwidth,keepaspectratio]{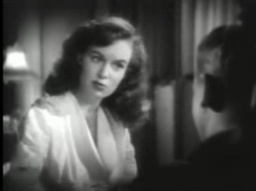}}
	\subfloat[] {\includegraphics[width= 0.110\columnwidth,keepaspectratio]{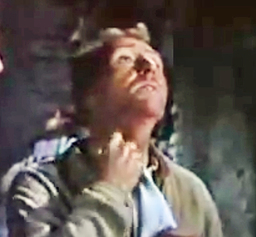}}
	\subfloat[] {\includegraphics[width= 0.110\columnwidth,keepaspectratio]{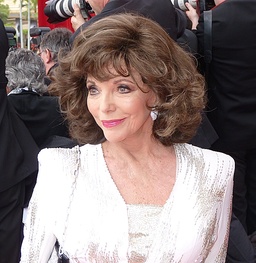}}
	\subfloat[] {\includegraphics[width= 0.110\columnwidth,keepaspectratio]{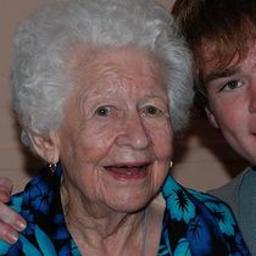}}
	\\ 
	\vspace{-22pt}
	\subfloat[{}] {{Align}}\hspace{0.4pt}
	\centering
	\subfloat[][{\hspace{-45pt}\tiny{Apparent~~~~1.03}\\\hspace{-45pt}\tiny{Estimation~~\color{red}1.32}}] 
       {\includegraphics[width= 0.110\columnwidth,keepaspectratio]{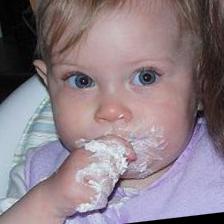}}
	\centering
	\subfloat[][\tiny{3.38}\\\color{red}\tiny{2.63}] {\includegraphics[width= 0.110\columnwidth,keepaspectratio]{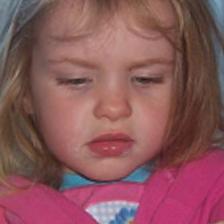}}
	\subfloat[][\tiny{6.34}\\\color{red}\tiny{5.17}] {\includegraphics[width= 0.110\columnwidth,keepaspectratio]{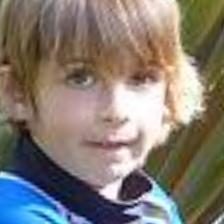}}
	\subfloat[][\tiny{18.44}\\\color{red}\tiny{17.72}] {\includegraphics[width= 0.110\columnwidth,keepaspectratio]{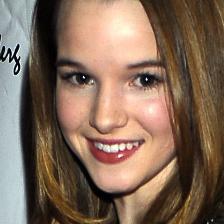}}
	\subfloat[][\tiny{24.23}\\\color{red}\tiny{22.93}] {\includegraphics[width= 0.110\columnwidth,keepaspectratio]{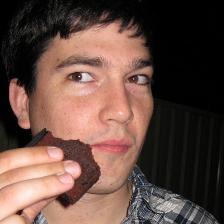}}
	\subfloat[][\tiny{28.74}\\\color{red}\tiny{28.51}] {\includegraphics[width= 0.110\columnwidth,keepaspectratio]{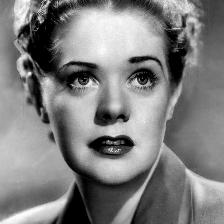}}
	\subfloat[][\tiny{37.15}\\\color{red}\tiny{36.32}] {\includegraphics[width= 0.110\columnwidth,keepaspectratio]{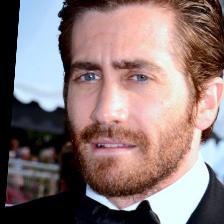}}
	\subfloat[][\tiny{51.86}\\\color{red}\tiny{52.40}] {\includegraphics[width= 0.110\columnwidth,keepaspectratio]{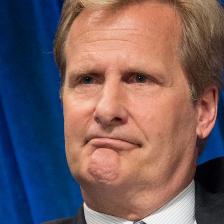}}
	\subfloat[][\tiny{69.00}\\\color{red}\tiny{67.62}] {\includegraphics[width= 0.110\columnwidth,keepaspectratio]{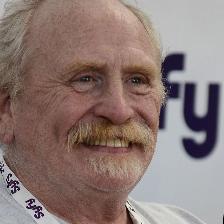}}
	\subfloat[][\tiny{79.86}\\\color{red}\tiny{78.43}] {\includegraphics[width= 0.110\columnwidth,keepaspectratio]{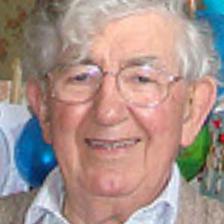}}
	\quad \vrule \quad
	\subfloat[][\tiny{11.24\\\color{blue}20.47}] {\includegraphics[width= 0.110\columnwidth,keepaspectratio]{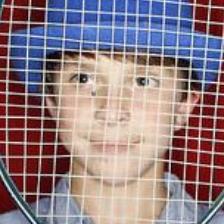}}
	\subfloat[][\tiny{29.81\\\color{blue}22.84}] {\includegraphics[width= 0.110\columnwidth,keepaspectratio]{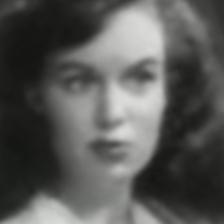}}
	\subfloat[][\tiny{41.71\\\color{blue}34.20}] {\includegraphics[width= 0.110\columnwidth,keepaspectratio]{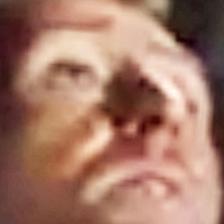}}
	\subfloat[][\tiny{66.84\\\color{blue}60.17}] {\includegraphics[width= 0.110\columnwidth,keepaspectratio]{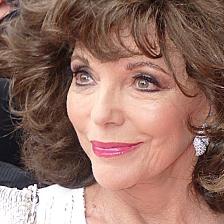}}
	\subfloat[][\tiny{88.50\\\color{blue}80.38}] {\includegraphics[width= 0.110\columnwidth,keepaspectratio]{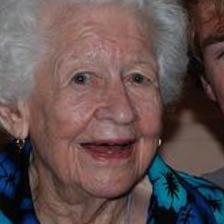}}\\
	\caption{Examples of apparent age estimation using DLDL-v2~(ThinAgeNet) on ChaLearn16 testing images. The left ten columns show good age estimations and the right five columns are poor cases.}\label{fig:age}
	\vspace{-10pt}
\end{figure*}

{In addition, we compare ThinAgeNet and TinyAgeNet with some recent network architectures such as ResNet-50~\cite{gustafsson2020energy,berg2021deep,han2015demographic}, ResNet-34~\cite{deng2021pml}, and MobileNet-v2~\cite{savchenko2019efficient}.}
{Note that we compute inference time with two different batch sizes ($n$=128 and $n$=1).  It has been observed in practice that when using a larger batch there is a significant speed up because of parallel computing characteristics of GPU. We can see that the TinyAgeNet is the best one among all networks both memory storage overhead and actual inference time. It is worth noting that although MobileNet-v2 has a lower theoretical computational complexity, it is still slower than our TinyAgeNet in terms of actual inference time. The main reason is that the depth-wise separable convolution in MobileNet-v2 requires efficient convolution computation, which may be still not well supported.}

\subsubsection{Visual Assessment}
Fig.~\ref{fig:age} shows some examples on ChaLearn16 testing images using our DLDL-v2 ThinAgeNet. In many cases, our solution is able to predict the age of faces accurately. Failures may come from some special cases such as occlusion, low resolution, heavy makeup and extreme pose.

\subsection{Experiments on Attractiveness Estimation}
\subsubsection{Attractiveness Estimation Datasets}
To further demonstrate the effectiveness of the proposed DLDL-v2, we perform extensive experiments on two facial attractiveness datasets: SCUT-FBP~\cite{xie2015scut} and CFD~\cite{ma2015chicago}.

SCUT-FBP~\cite{xie2015scut} is a widely used facial beauty assessment dataset. It contains 500 Asian female faces with neutral expressions, simple backgrounds, no accessories, and minimal occlusion. Each face is scored by 75 workers with a 5-point scale, in which 1 means strong agreement about the face being the least attractive and 5 means strong agreement about the face being the most attractive. For each face, its mean score and the corresponding standard deviation are given. We follow the setting in~\cite{fan2017label} and ~\cite{xie2015scut}, 80\% images are randomly selected as the training set, and the remain 20\% as the testing set.

CFD~\cite{ma2015chicago} provides high-resolution and standardized photographs with meaningful annotations~(\eg, attractiveness, babyfacedness and expression \etc). Unlike SCUT-FBP, this dataset includes male and female faces of multiple ethnicity~(Asian, Black, Latino, and White) between the ages of 17-65. Similar to SCUT-FBP, each faces is scored by some participants with diverse background in a 7-point scale~(1 = Not at all, 7 = Extremely). In this study, we employ all 597 faces with natural expression and the corresponding attractiveness scores for experiments. We use 80\% images for training and the remain 20\% for testing.

\begin{table}
	\centering
	\small
	\caption{Comparisons with state-of-the-art methods for facial attractiveness estimation on SCUT-FBP dataset. \textsuperscript{\dag}indicates methods using ten-fold cross validation, 90$\%$ of images for training and 10$\%$ for evaluation and \textsuperscript{\ddag}indicates methods using multi-features fusion.}\label{tab:sotascut}
	  \scalebox{0.9}{
	\begin{tabular}{|@{\;}l@{\;}| @{\;}r@{\;}| @{}c@{\;} | *{3}{@{\;}c@{\;}}|}
		\hline
		Methods &{Venue Year}   &{Model}  &MAE$\downarrow$ &RMSE$\downarrow$ &PC$\uparrow$    \\
		
		\hline
		Regression~\cite{xie2015scut}                             &SMC 2015   &G+Tfeats     &0.393   &0.515   &0.648    \\
		SLDL~\cite{rensense}\textsuperscript{\dag}        &IJCAI 2017     &LBP+Hog+Gabor  &0.302  &0.408    &-           \\
		CNN~\cite{xie2015scut}~                                     &SMC 2015    &Six-layer CNN     &-          &-           &0.819   \\
		LDL~\cite{fan2017label}~                                     &TMM 2017    &ResNet-50   &0.217  &0.300    &0.917   \\           		
		LDL~\cite{fan2017label}\textsuperscript{\ddag}   &TMM 2017   &ResNet-50+GFeats   &0.213   &0.278    &\textbf{0.930}  \\
		R2-ResNeXt~\cite{lin2018r}                                &ICPR 2018   &ResNeXt    &0.242   &0.305    &0.896  \\
		DALDL~\cite{chen2019facial}                              &CCBR 2019 &VGG-16    &0.227    &0.312    &0.903 \\
                 MT-ResNet~\cite{xu2021mt}                                &CDS 2021    &ResNet-50    &0.246   &0.321    &0.891  \\
		\hline
		\multicolumn{2}{|c|} {DLDL-v2~(TinyAttNet)}     &Tiny VGG-16 &0.221 &0.294  &0.915   \\
	        \multicolumn{2}{|c|} {DLDL-v2~(ThinAttNet)}     &Thin VGG-16  &\textbf{0.212} &\textbf{0.273}  &\textbf{0.930}  \\
		\hline
	\end{tabular}}
\end{table}

\subsubsection{Attractiveness Estimation Results}
In Table~\ref{tab:sotascut} and \ref{tab:sotacfd} , we report the performance on SCUT-FBP and CFD and compare with the state-of-the-art methods in the literature. 

\begin{table}
	\centering
	\small
	\caption{Comparisons with state-of-the-art methods for facial attractiveness estimation on CFD dataset. }\label{tab:sotacfd}
	\begin{tabular}{|@{\;}l@{\;}| *{3}{@{\;}c@{\;}}|}
		\hline
		Methods     &MAE$\downarrow$   &RMSE$\downarrow$  & PC$\uparrow$\\
		\hline
		DLDL-v2~(TinyAttNet)                &0.400 &0.521 &0.716\\
	        DLDL-v2~(ThinAttNet)              &\textbf{0.364} &\textbf{0.472} &\textbf{0.766}\\
		\hline
	\end{tabular}
\end{table}

Comparing with those methods using hand-crafted features, such as Regression~\cite{xie2015scut} and SLDL~\cite{rensense}, the proposed DLDL-v2~(ThinAttNet) achieves 0.930 PC and 0.212 MAE on SCUT-FBP. It outperforms Regression~\cite{xie2015scut} by 0.282 in PC, and improves SLDL~\cite{rensense} by 0.135 in RMSE. What is more, for those methods using deep label distribution, such as LDL (ResNet50)~\cite{fan2017label} as one of the state-of-the-art methods, our DLDL-v2 still outperforms it. Furthermore, our method is comparable to the fusional solution of deep features and geometric features in~\cite{fan2017label}. There are two major reasons. First, our pre-trained model is trained on a face recognition dataset which is closer to facial attractiveness than those object classification datasets~(ResNet50 is trained by ImageNet) in~\cite{fan2017label}. Second, we jointly learn label distribution and regress the facial attractiveness score in DLDL-v2, which can effectively erase the inconsistency between training objective and evaluation metric~(MAE).

\begin{figure*}
	\captionsetup[subfigure]{labelformat=empty}
	\captionsetup[subfigure]{justification=centering}
	\centering
	\subfloat[{}] {\vspace{-20pt}\footnotesize Input\;\;}\hspace{2pt}
	\subfloat[] {\includegraphics[width= 0.110\columnwidth,keepaspectratio]{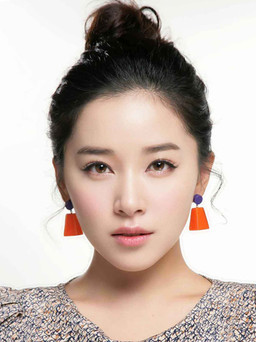}}
	\subfloat[] {\includegraphics[width= 0.110\columnwidth,keepaspectratio]{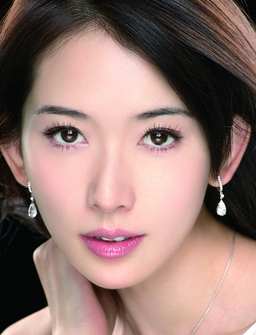}}
	\subfloat[] {\includegraphics[width= 0.110\columnwidth,keepaspectratio]{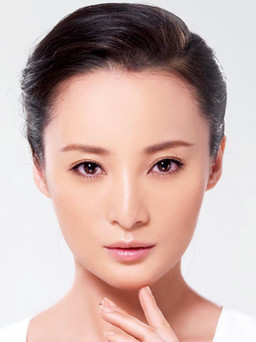}}
	\subfloat[] {\includegraphics[width= 0.110\columnwidth,keepaspectratio]{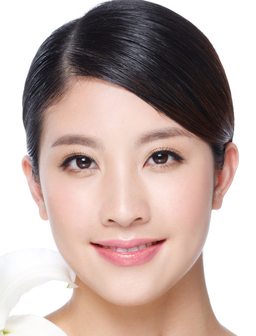}}
	\subfloat[] {\includegraphics[width= 0.110\columnwidth,keepaspectratio]{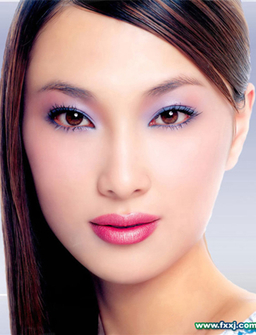}}
	\subfloat[] {\includegraphics[width= 0.110\columnwidth,keepaspectratio]{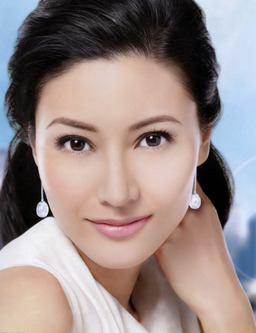}}
	\subfloat[] {\includegraphics[width= 0.110\columnwidth,keepaspectratio]{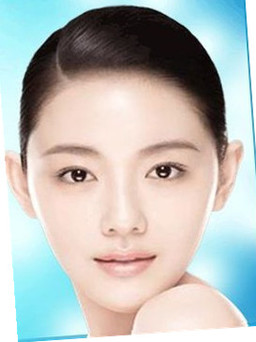}}
	\subfloat[] {\includegraphics[width= 0.110\columnwidth,keepaspectratio]{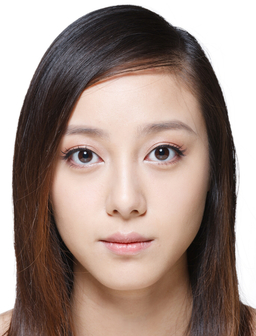}}
	\quad  \quad
	\subfloat[] {\includegraphics[width= 0.110\columnwidth,keepaspectratio]{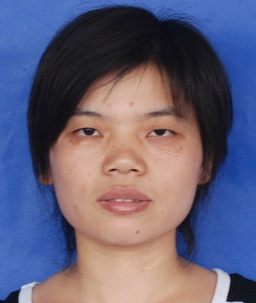}}
	\subfloat[] {\includegraphics[width= 0.110\columnwidth,keepaspectratio]{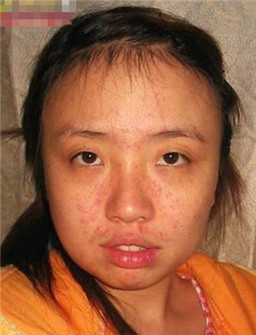}}
	\subfloat[] {\includegraphics[width= 0.110\columnwidth,keepaspectratio]{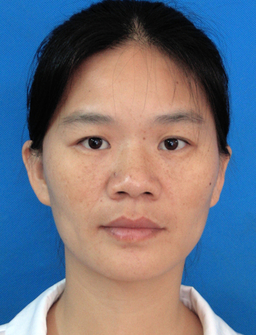}}
	\subfloat[] {\includegraphics[width= 0.110\columnwidth,keepaspectratio]{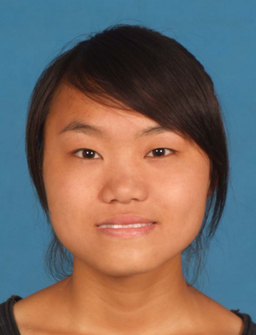}}
	\subfloat[] {\includegraphics[width= 0.110\columnwidth,keepaspectratio]{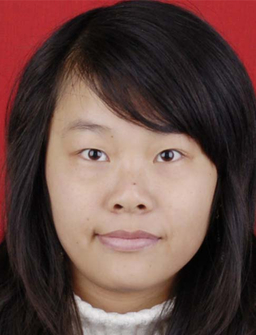}}
	\subfloat[] {\includegraphics[width= 0.110\columnwidth,keepaspectratio]{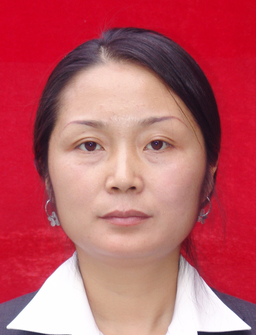}}
	\subfloat[] {\includegraphics[width= 0.110\columnwidth,keepaspectratio]{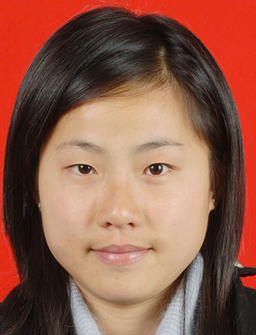}}
	\subfloat[] {\includegraphics[width= 0.110\columnwidth,keepaspectratio]{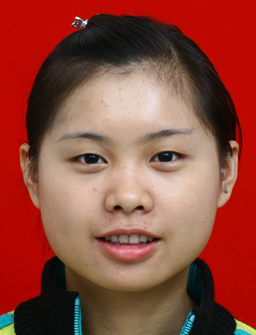}}
	\\\vspace{-22pt}
	\subfloat[{}] {\footnotesize Align\;\;} \hspace{0.4pt}
	\subfloat[][\hspace{-35pt}Human~~~4.67\\\hspace{-45pt}Estimation~~\color{red}{4.47}] 
       {\includegraphics[width= 0.110\columnwidth,keepaspectratio]{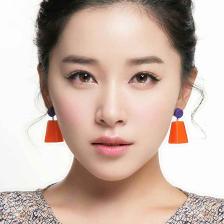}}
	\subfloat[][4.66\\\color{red}{4.40}] {\includegraphics[width= 0.110\columnwidth,keepaspectratio]{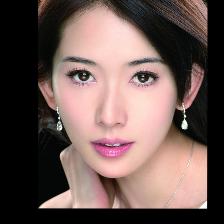}}
	\subfloat[][4.38\\\color{red}{4.39}] {\includegraphics[width= 0.110\columnwidth,keepaspectratio]{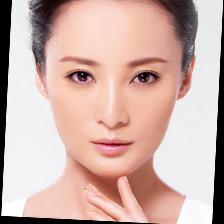}}
	\subfloat[][4.08\\\color{red}{4.32}] {\includegraphics[width= 0.110\columnwidth,keepaspectratio]{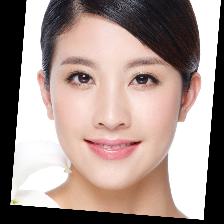}}
	\subfloat[][3.73\\\color{blue}{4.31}] {\includegraphics[width= 0.110\columnwidth,keepaspectratio]{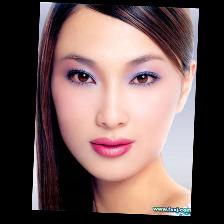}}
	\subfloat[][4.31\\\color{red}{4.29}] {\includegraphics[width= 0.110\columnwidth,keepaspectratio]{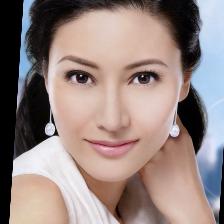}}
	\subfloat[][4.50\\\color{red}{4.22}] {\includegraphics[width= 0.110\columnwidth,keepaspectratio]{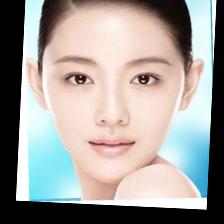}}
	\subfloat[][4.17\\\color{red}{4.12}] {\includegraphics[width= 0.110\columnwidth,keepaspectratio]{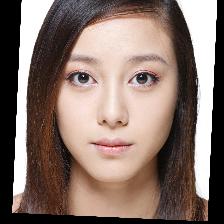}}
	\quad  \vrule  \quad
	\subfloat[][1.59\\\color{red}{1.68}] {\includegraphics[width= 0.110\columnwidth,keepaspectratio]{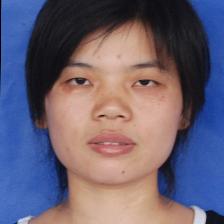}}
	\subfloat[][1.38\\\color{blue}{1.82}] {\includegraphics[width= 0.110\columnwidth,keepaspectratio]{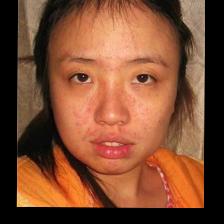}}
	\subfloat[][2.01\\\color{red}{1.87}] {\includegraphics[width= 0.110\columnwidth,keepaspectratio]{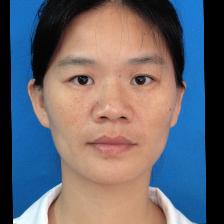}}
	\subfloat[][2.00\\\color{red}{1.90}]{\includegraphics[width= 0.110\columnwidth,keepaspectratio]{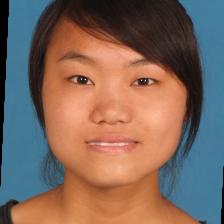}}
	\subfloat[][2.03\\\color{red}{1.93}] {\includegraphics[width= 0.110\columnwidth,keepaspectratio]{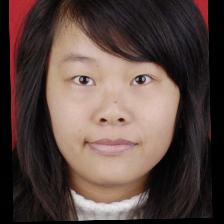}}
	\subfloat[][2.28\\\color{blue}{1.94}] {\includegraphics[width= 0.110\columnwidth,keepaspectratio]{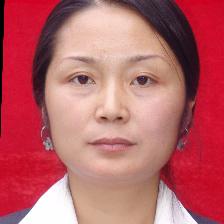}}
	\subfloat[][2.27\\\color{blue}{1.94}] {\includegraphics[width= 0.110\columnwidth,keepaspectratio]{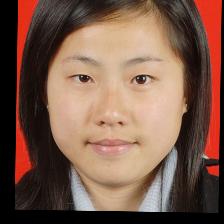}}
	\subfloat[][2.04\\\color{red}{1.95}] {\includegraphics[width= 0.110\columnwidth,keepaspectratio]{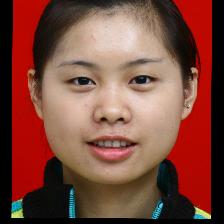}}\\
	\caption{Examples of facial attractiveness estimation using DLDL-v2~(ThinAttNet) on SCUT-FBP testing images. These images are ranked by our DLDL-v2 prediction scores. The top eight faces are showed in the left side, and the bottom eight faces are showed in the right side. The {\color{red}{red}} numbers denote good estimation~(MAE is less than 0.3), and the {\color{blue}{blue}} numbers denote poor estimation (MAE is greater than 0.3).}\label{fig:att}
	\vspace{-10pt}
\end{figure*}

\subsubsection{Visual Assessment}
In order to intuitively visualize the prediction performance of our DLDL-v2 on facial attractiveness task, we show the top eight and bottom eight test images based on the prediction scores of DLDL-v2 with ThinAttNet in Fig.~\ref{fig:att}. On selected 16 testing images, prediction scores of 12 images highly match with those of human raters. This result qualitatively demonstrates that our DLDL-v2 is able to generate human-like results. In addition, some possible facial attractiveness cues may be observed via comparing between the top and bottom faces with attractiveness score. Generally speaking, faces with higher attractive scores have smoother and lighter skin, oval face with larger eyes, narrower nose with a pointed tip, and better harmony in facial organs than those with lower scores.

\subsection{Ablation Study and Diagnostic Experiments}

DLDL-v2~(ThinAgeNet) is employed for ablation study on facial age datasets in this section. We firstly investigate the efficacy of the proposed data augmentation and the pooling strategy. For fair comparison, we fix $\bigtriangleup l =1$ and $\lambda=1$. Then, to investigate the effectiveness of the proposed joint learning mechanism, we compare it with two stage and single stage methods under the same setting. At last, we also explore the sensitivity of hyper-parameters in our DLDL-v2. 

\subsubsection{Influence of Data Augmentation} 
Data augmentation techniques increase the amount of training data using information only in training set, which is an effective way to reduce the over-fitting of deep models. From Table~\ref{tab:as}, we can observe 0.26$\sim$0.27 MAE improvements on apparent age datasets and 0.38 MAE improvement on Morph using data augmentation. This indicates that data augmentation can greatly improve the performance of age estimation.

\begin{table}
 \centering
 \small
 \caption{Comparison of different methods for age estimation.}\label{tab:as}
 \begin{tabular}{|@{\;}c@{\;} @{\;}c@{\;}| *{2}{@{\;}c@{\;}}| *{2}{@{\;}c@{\;}}| @{\;}c@{\;}|}
  \hline
    \multicolumn{2}{|c}{Factors} &\multicolumn{2}{c}{ChaLearn15}   &\multicolumn{2}{c}{ChaLearn16} &{Morph}\\
  \cline{1-7}
   Aug &Pooling        & MAE &$\epsilon$-error &MAE &$\epsilon$-error & MAE\\
  \hline 
    $\times$ &HP   &3.399  &0.303   &3.717  &0.290  &2.346\\
    $\surd$  &GAP   &3.210  &0.282   &3.539  &0.274  &2.039\\
    $\surd$  &HP   &\textbf{3.135} &\textbf{0.272}  &\textbf{3.452} &\textbf{0.267} &\textbf{1.969}\\
  \hline
 \end{tabular}
\end{table}

\subsubsection{Effectiveness of Pooling Strategy}
 GAP is  one of the most popular and simple method for aggregating the spatial information of feature maps in state-of-the-art network architecture such as ResNet~\cite{he2016deep}.  It outputs the spatial average of each feature map of the last convolution layer. Max-pooling takes the maximal value of each small region (\eg, $2\times2$) in a feature map as its output. HP is constructed by a max-pooling and a GAP layer. HP firstly encourages the network to learn a discriminative feature in a small region via max-pooling, then all discriminative features are aggregated by GAP. Thus, the feature of HP is more discriminative than that of GAP. If we directly use global max-pooling instead of HP, the training of network easily fall into over-fitting. To explore the effect of the pooling strategy, we further use the HP to replace the traditional GAP when combining data augmentation. It can be seen in Table~\ref{tab:as} that the proposed HP can consistently reduce the prediction error on all datasets.

\subsubsection{Comparisons with Two Stage Methods}
We compare the proposed approach with two stage methods considering two types of features. The first one is the BIF~\cite{guo2009human}, as the most successful hand-crafted age feature, which was widely used in age estimation. The second one is CNN features which are extracted from our pre-trained face recognition model. For BIF, we adopt 6 bands and 8 orientations~\cite{guo2009human}, which produces 4616-dimensional features. The CNN features are extracted from the hybrid pooling layer of the pre-trained model and their dimension is 256. These features are normalized by $\ell_2$ without using any dimensionality reduction technique.

We choose three classical age estimation algorithms, including SVR~\cite{guo2009human}, OHRank~\cite{chang2011ordinal} and BFGS-LDL~\cite{geng2013facial}. For SVR and OHRank, the Liblinear software is used to train regression or classification models.\footnote{\url {https://www.csie.ntu.edu.tw/~cjlin/liblinear/}} For BFGS-LDL, we use the open source LDL package.\footnote{\url{http://ldl.herokuapp.com/download}} Instead of age prediction with the maximal probability in~\cite{geng2013facial}, we use the expected value over prediction distribution because it has better performance.

\begin{table}
 \centering
 \small
 \caption{Comparisons with two stage methods for age estimation~(lower is better).
 }\label{tab:ctage}
 \begin{tabular}{
 |@{\;}l@{\;}| @{\;}c@{\;}|@{}c@{\;} @{\;}c@{\;}| *{2}{@{\;}c@{\;}}| @{\;}r@{\;}| }
  \hline
    \multirow{2}*{Feats}      &\multirow{2}*{Methods} &\multicolumn{2}{c}{ChaLearn15}   &\multicolumn{2}{c}{ChaLearn16} &{Morph}\\\cline{3-7}
    &       &~MAE &$\epsilon$-error &MAE &$\epsilon$-error & MAE\\
  \hline
 \multirow{3}*{BIF} &SVR           &~6.832 &0.545   &9.225 &0.595 &4.303\\
  &OHRank         &~6.403 &0.525   &7.680 &0.533 &3.841\\
  &BFGS-LDL       &~6.441 &0.505   &7.626 &0.515 &3.883\\
   \hline
  \multirow{3}*{CNN} &SVR           &~5.333 &0.471   &6.348 &0.495 &4.370\\
  &OHRank         &~4.202 &0.383   &4.668 &0.380 &3.919\\
  &BFGS-LDL       &~4.037 &0.359   &4.457 &0.345 &3.865\\
  \hline
   \multicolumn{2}{|c|}{DLDL-v2}  
      &~\textbf{3.135} &\textbf{0.272}  &\textbf{3.452} &\textbf{0.267} &\textbf{1.969}\\
  \hline
 \end{tabular}
\end{table}

The experimental results are shown in Table~\ref{tab:ctage}. First, OHRank and BFGS-LDL using BIF and CNN features have similar performances on all datasets. This further validates our previous analysis that ranking is learning label distribution. Second, our proposed approach significantly outperforms all baseline methods. The major reason is that two stage methods cannot learn visual representations. This suggests that it is crucially important to jointly learn visual features and recognition model using an end-to-end manner. At last, OHRank and BFGS-LDL are much better than SVR, which indicates learning label distribution can really help us to improve estimation performance.
\subsubsection{Comparisons with Single Stage Methods}
We employ six very strong methods under the same setting as baselines: 

\squishlist
 \item \textbf{MR}: In MR, the ground-truth label $\mathrm y$ is projected to $[-1,1]$ by a linear transform. For MR, we need to make a little modification to DLDL-v2. Specifically, we add an FC layer with single output after HP, and follow a hyperbolic tangent activation function $f(x) = \tanh(x)$ for speedup convergence. The $\ell_2$ and $\ell_1$ loss function is used to train MR. 
 
 \item \textbf{DEX}: In DEX, true label $\mathrm y$ is quantized to different label group, which is treated as a class. To train DEX, we only need remove the expectation module and modify loss function to cross-entropy loss in DLDL-v2. In inference time, an expected value over prediction probabilities is used for final estimation.

 \item \textbf{Ranking}: In~\cite{chen2017using,Chen2017Deep}, multiple binary classification networks are independently trained, which lead to time-consuming of training and storage overhead of model. We \emph{propose} a new multiple output CNN and jointly train these binary classifiers. Specifically, we firstly remove the label distribution and expectation module in DLDL-v2. Then, we add an FC layer with $K-1$ output units and follow a sigmoid layer. For training Ranking CNN, we employ $K-1$ binary cross-entropy loss. In inference stage, the prediction is computed by $\hat y = l_{i^*}$, where $i^*=1+\sum_{k=1}^{K-1}[\delta(x)>0.5]$. $[\cdot]$ denotes the truth-test operator, which is 1 if the inner condition is true, and 0 otherwise. Our experiments showed that this new setup has lower MAE than that in~\cite{niu2016ordinal,chen2017using,Chen2017Deep}.
 \item \textbf{ER~($\ell_1$)}: We only employ the expectation regression~(ER) loss $L_{er}$ to optimize DLDL-v2's parameters via removing label distribution loss $L_{ld}$ in Eq.~\eqref{eq:loss1}.
 \item \textbf{DLDL}: We set $\lambda=0$ in Eq.~\eqref{eq:loss1} to learn DLDL-v2.
\squishend

Table~\ref{tab:csage} reports the results of all single stage methods. We can see that the MAE and $\epsilon$-error of Ranking, ER and DLDL methods are significantly lower than that of MR and DEX on all datasets. This indicates that utilizing label distribution is helpful to reduce age estimation error. Meanwhile, we also find that the prediction error of Ranking is close to that of DLDL, which conforms to the analysis in Section~\ref{subsec:rem}. Furthermore, the performance of DLDL is better than that of Ranking, which suggests that learning p.d.f. is more effective than learning c.d.f. It is noteworthy that ER~($\ell_1$) and DLDL are two extreme cases of our DLDL-v2. DLDL-v2 consistently outperforms ER~($\ell_1$) and DLDL on all datasets, which indicates the joint learning can ease the difficult of network optimization. In Table~\ref{tab:csage}, we can see that the proposed joint learning achieves the best performance among all methods. It means that erasing the inconsistency between training and evaluation stages can help us make a better prediction.

\begin{table}
 \centering
 \small
 \caption{Comparisons with single stage methods for age estimation~(lower is better).
 }\label{tab:csage}
 \begin{tabular}{|@{\;}l@{\;}|@{}c@{\;}@{\;}c@{\;}| *{2}{@{\;}c@{\;}}| @{\;}c@{\;}|}
  \hline
  \multirow{2}*{Methods}   &\multicolumn{2}{c}{ChaLearn15}   &\multicolumn{2}{c}{ChaLearn16} &{Morph}\\
  \cline{2-6}
                          &~MAE &$\epsilon$-error &MAE &$\epsilon$-error & MAE\\
  \hline
  MR~($\ell_2$) &~3.665 &0.337   &3.696 &0.294 &2.282\\
  MR~($\ell_1$) &~3.655 &0.334   &3.722 &0.301 &2.347\\
  DEX           &~3.558 &0.306  &4.163 &0.332 &2.311\\
  Ranking       &~3.365 &0.298  &3.645 &0.290 &2.164\\
  \hline
  ER~($\ell_1$) &~3.287 &0.291  &3.641 &0.282 &2.214\\
  DLDL          &~3.228 &0.285  &3.509 &0.272 &2.132\\ 
  \hline
  {DLDL-v2}    &~\textbf{3.135} &\textbf{0.272}  &\textbf{3.452} &\textbf{0.267} &\textbf{1.969}\\
  \hline
 \end{tabular}
\end{table}
\subsubsection{Sensitivity of Hyper-parameters}

\begin{table}
 \centering
 \small
 \caption{The influences of hyper-parameters for our DLDL-v2. 
 }\label{tab:hp}
 \begin{tabular}{|*{2}{@{\;}c@{\;}}| *{2}{@{\;}c@{\;}}| *{2}{@{\;}c@{\;}}|  @{\;}r@{\;}| }
  \hline
    \multicolumn{2}{|c}{Hyper-param} &\multicolumn{2}{c}{ChaLearn15}   &\multicolumn{2}{c}{ChaLearn16} &{Morph}\\\cline{1-7}
    $\lambda$ &$\bigtriangleup l$~($K$)        & MAE &$\epsilon$-error &MAE &$\epsilon$-error & MAE\\
  \hline   
    0.01  &1~(101)   &3.223  &0.282   &3.493  &0.270  &\textbf{1.960}\\
    0.10  &1~(101)  &3.188  &0.278   &3.455  &0.268  &1.972\\
    1.00  &1~(101)  &\textbf{3.135} &\textbf{0.272}  &\textbf{3.452} &0.267 &1.969\\
    10.00 &1~(101)  &3.144 &0.273  &3.487  &0.270 &1.977\\
    \hline
    1.00  &4~(26)   &3.182  &0.276   &3.473  &0.270  &1.963\\
    1.00  &2~(51)   &3.184  &0.274   &3.484  &0.271  &1.963\\
    1.00  &0.50~(201)   &3.184  &0.278   &3.484  &0.269  &1.992\\  
    1.00  &0.25~(401)   &3.167  &0.274  &3.459  &\textbf{0.265}    &2.028\\     
  \hline
 \end{tabular}
\end{table}

We explore the influence of hyper-parameters $\lambda$ and $\bigtriangleup l$, where $\lambda$ is a weight which balances the importance between label distribution and expectation regression loss, and $\bigtriangleup l$ refers to the number of discrete labels ($K$). In Table~\ref{tab:hp}, we report results on all three age datasets with different value of $\lambda$ and $\bigtriangleup l$. We can see that our method is not sensitive to $\lambda$ and $\bigtriangleup l$ with $0.01 \leq \lambda \leq 10$ and $0.25 \leq \bigtriangleup l \leq 4$. Note that, too many discrete labels lead to little training samples for per class in DEX~\cite{rothe2016deep}, which may make prediction less precise. However, our method can ease the problem,
because the training samples associated with each class label is significantly increased without actually increase the number of the total training examples. Surprising, there is also a good enough performance when the number output neurons~(\ie, $K$) is 26. In our experiment, we fixed hyper-parameters $\bigtriangleup l =1$ and
$\lambda=1$ without carefully tuning them. In practice, it is convenient to find optimal hyper-parameters using a hold-out set.

\section{Understanding DLDL-v2}\label{sec:ud}
We have demonstrated that DLDL-v2 has excellent performance for facial age and attractiveness estimation. A natural problem is how DLDL-v2 makes the final decision for an input facial image. In this section, we try to answer this question. Then, we analyze why it can work well when compared with existing methods.

\subsection{How Does DLDL-v2 Estimate Facial Attributes?}
In order to understand how DLDL-v2 makes the final decision for an input facial image, we visualize a score map that can intuitively show which regions of face image are related to the network decision. To obtain the score map, we firstly employ a class-discriminative localization technique~\cite{zhou2016learning} that can generate class activation maps. Then, these activation maps are aggregated by predicted probabilities. 

Let us briefly review our framework. The last convolution block produce activation maps $\mathbf F^j\in \mathcal R^{h\prime \times w\prime}$. These activations are then spatially pooled by a hybrid pooling and linearly transformed~(\ie, Eq.~\eqref{eq:linear}) to produce probabilities $\vec {\hat p}\in \mathcal R^K$ with a label distribution module. To produce class activation maps, we apply linearly transform layer to $\mathbf F^{j}$ as follows
\begin{equation}\label{eq:classmap}
\mathbf A^k = \sum_j w_{kj}\mathbf F^{j} + b_k\,.
\end{equation}
Then, the score map can be derived by
\begin{equation}\label{eq:score}
\mathbf S = \sum_k \hat {p_k} \mathbf A^k\,.
\end{equation}
In Eq.~\eqref{eq:score}, the value of $S_{ij}$ represents the contribution of the network's decision at position of $i$-th row and $j$-th column. Bigger values mean more contributions and vice versa.  For comparing the correspondence between highlighted regions in $\mathbf S$ and an input image, we scale $\mathbf S$ to the size of an input image. 

\begin{figure}
    \captionsetup[subfigure]{labelformat=empty}
    \centering
    \subfloat[] {\includegraphics[width= 0.1\columnwidth]{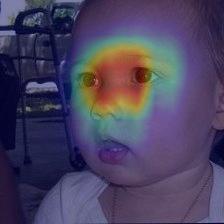}}
    \subfloat[] {\includegraphics[width= 0.1\columnwidth]{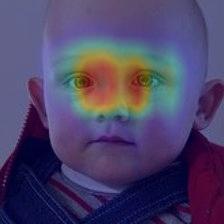}}
    \subfloat[] {\includegraphics[width= 0.1\columnwidth]{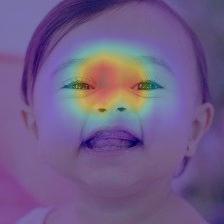}}
    \subfloat[] {\includegraphics[width= 0.1\columnwidth]{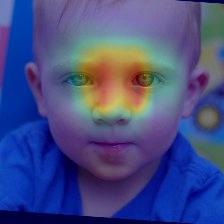}}
    \subfloat[] {\includegraphics[width= 0.1\columnwidth]{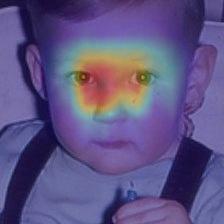}}
    \subfloat[] {\includegraphics[width= 0.1\columnwidth]{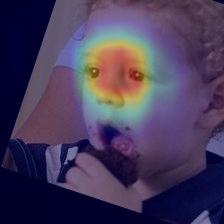}}
    \subfloat[] {\includegraphics[width= 0.1\columnwidth]{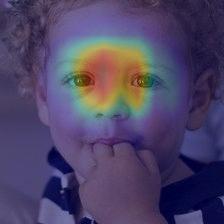}}
    \subfloat[] {\includegraphics[width= 0.1\columnwidth]{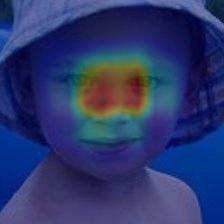}}
    \subfloat[] {\includegraphics[width= 0.1\columnwidth]{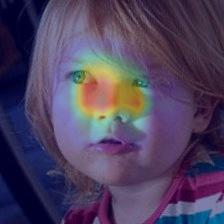}}
    \subfloat[] {\includegraphics[width= 0.1\columnwidth]{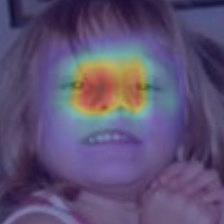}}
    \\ \vspace{-25pt}
    \subfloat[] {\includegraphics[width= 0.1\columnwidth]{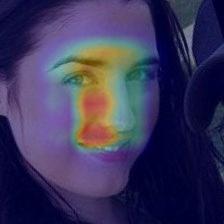}}
    \subfloat[] {\includegraphics[width= 0.1\columnwidth]{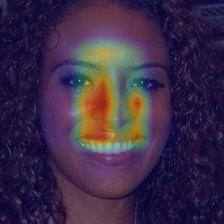}}
    \subfloat[] {\includegraphics[width= 0.1\columnwidth]{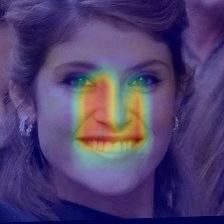}}
    \subfloat[] {\includegraphics[width= 0.1\columnwidth]{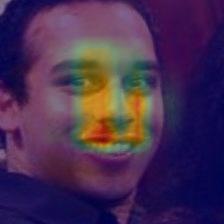}}
    \subfloat[] {\includegraphics[width= 0.1\columnwidth]{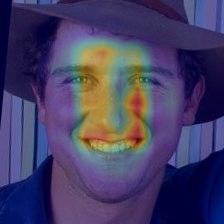}}
    \subfloat[] {\includegraphics[width= 0.1\columnwidth]{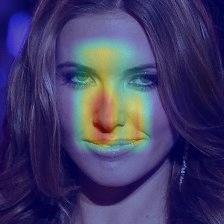}}
    \subfloat[] {\includegraphics[width= 0.1\columnwidth]{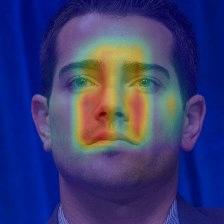}}
    \subfloat[] {\includegraphics[width= 0.1\columnwidth]{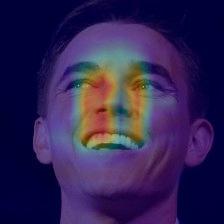}}
    \subfloat[] {\includegraphics[width= 0.1\columnwidth]{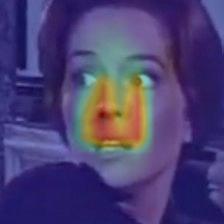}}
    \subfloat[] {\includegraphics[width= 0.1\columnwidth]{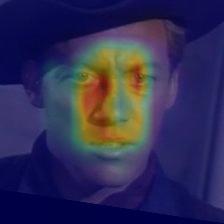}}
     \\ \vspace{-25pt}
    \subfloat[] {\includegraphics[width= 0.1\columnwidth]{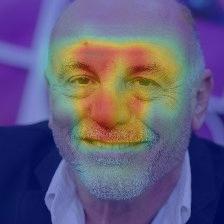}}
    \subfloat[] {\includegraphics[width= 0.1\columnwidth]{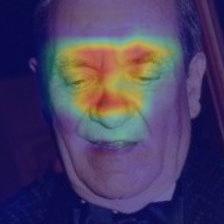}}
    \subfloat[] {\includegraphics[width= 0.1\columnwidth]{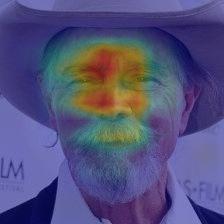}}
    \subfloat[] {\includegraphics[width= 0.1\columnwidth]{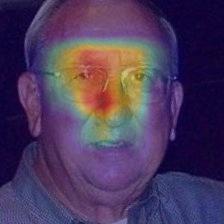}}
    \subfloat[] {\includegraphics[width= 0.1\columnwidth]{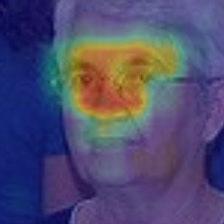}}
    \subfloat[] {\includegraphics[width= 0.1\columnwidth]{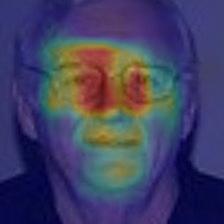}}
    \subfloat[] {\includegraphics[width= 0.1\columnwidth]{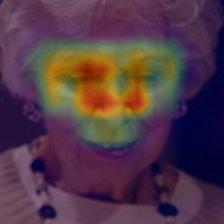}}
    \subfloat[] {\includegraphics[width= 0.1\columnwidth]{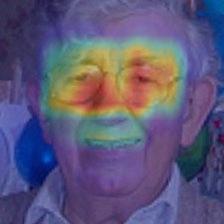}}
    \subfloat[] {\includegraphics[width= 0.1\columnwidth]{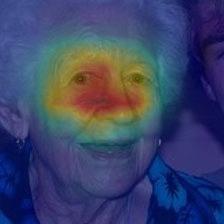}}
    \subfloat[] {\includegraphics[width= 0.1\columnwidth]{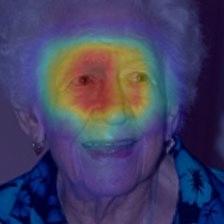}}
    \vspace{-10pt}
 \caption{Examples of the score map visualizations on ChaLearn16 testing images. The first row is for infants~($[0,3]$), the second row is for adults~($[20,35]$) and the third row is for senior people~($[65,100]$)~(Best viewed in color).}\label{fig:ageactvis}
  \vspace{-10pt}
\end{figure}

In Fig.~\ref{fig:ageactvis}, we visualize the score maps of testing images~(ChaLearn16) coming from different age group. we can see that the highlighted regions~(\ie, red regions) are significantly different for different age group faces. For infants, the highlighted region locates in the center of two eyes. For adults, the strong areas include two eyes, nose and mouth. For senior people, the highlighted regions consist of the forehead, brow, two eyes and nose. In short, the network uses different patterns to estimate different age. 

\begin{figure}
    \captionsetup[subfigure]{labelformat=empty}
    \centering
    \subfloat[]{\includegraphics[width= 0.1\columnwidth,keepaspectratio]{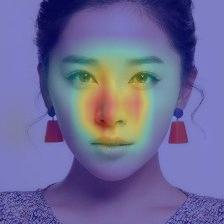}}
    \subfloat[]{\includegraphics[width= 0.1\columnwidth,keepaspectratio]{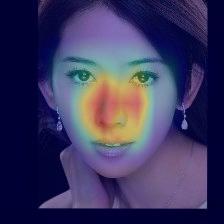}}
    \subfloat[]{\includegraphics[width= 0.1\columnwidth,keepaspectratio]{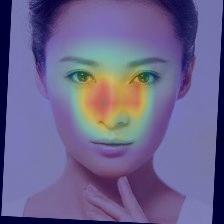}}
    \subfloat[]{\includegraphics[width= 0.1\columnwidth,keepaspectratio]{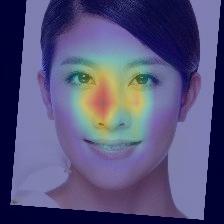}}
    \subfloat[]{\includegraphics[width= 0.1\columnwidth,keepaspectratio]{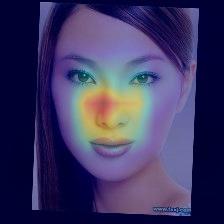}}
    \subfloat[]{\includegraphics[width= 0.1\columnwidth,keepaspectratio]{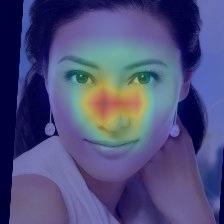}}
    \subfloat[]{\includegraphics[width= 0.1\columnwidth,keepaspectratio]{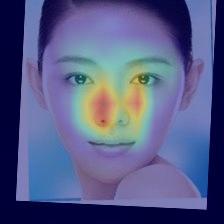}}
    \subfloat[]{\includegraphics[width= 0.1\columnwidth,keepaspectratio]{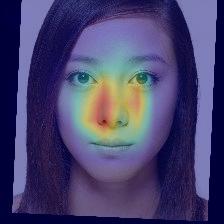}}
    \subfloat[]{\includegraphics[width= 0.1\columnwidth,keepaspectratio]{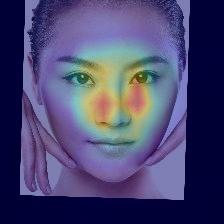}}  
    \subfloat[]{\includegraphics[width= 0.1\columnwidth,keepaspectratio]{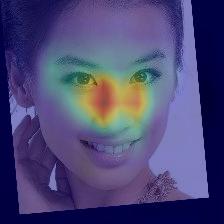}}  
    \\ \vspace{-25pt}
    \subfloat[]{\includegraphics[width= 0.1\columnwidth,keepaspectratio]{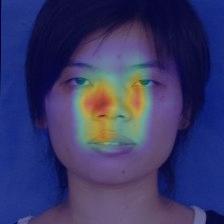}}
    \subfloat[]{\includegraphics[width= 0.1\columnwidth,keepaspectratio]{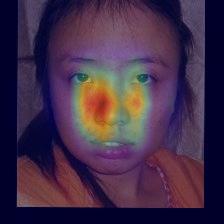}}
    \subfloat[]{\includegraphics[width= 0.1\columnwidth,keepaspectratio]{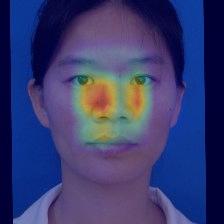}}
    \subfloat[]{\includegraphics[width= 0.1\columnwidth,keepaspectratio]{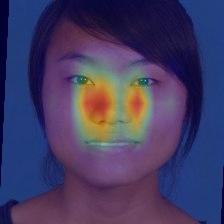}}
    \subfloat[]{\includegraphics[width= 0.1\columnwidth,keepaspectratio]{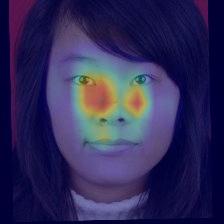}}
    \subfloat[]{\includegraphics[width= 0.1\columnwidth,keepaspectratio]{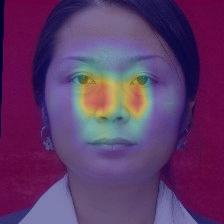}}
    \subfloat[]{\includegraphics[width= 0.1\columnwidth,keepaspectratio]{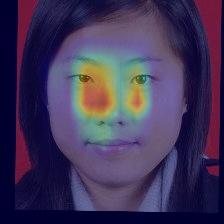}}
    \subfloat[]{\includegraphics[width= 0.1\columnwidth,keepaspectratio]{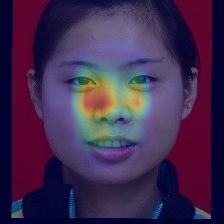}}
    \subfloat[]{\includegraphics[width= 0.1\columnwidth,keepaspectratio]{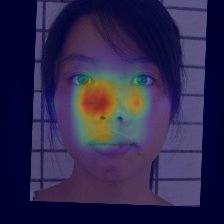}}
    \subfloat[]{\includegraphics[width= 0.1\columnwidth,keepaspectratio]{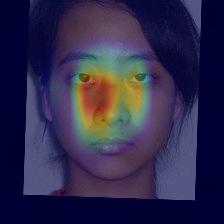}}
 \vspace{-10pt}
 \caption{Examples of the score map visualizations on SCUT-FBP testing images. The first and second row shows Top-10 and Bottom-10 faces rated by prediction scores~(Best viewed in color).}\label{fig:attactvis}
  \vspace{-10pt}
\end{figure}

We also show some examples coming from SCUT-FBP testing images in Fig.~\ref{fig:attactvis}. We can observe that it is not significant for the highlighted regions between these faces with higher attractiveness score and that of lower score. An explanation is that DLDL-v2 may be able to estimate facial attractiveness through simply comparing the difference of the common facial traits such as eyebrows, eyes, nose, mouth~\etc. In fact, the SCUT-FBP dataset indeed has the lower complexity~(female faces with simple backgrounds, no  accessories, and minimal occlusion) than age estimation on ChaLearn16.

\subsection{Sensitivity to Different Face Regions}
To further quantitatively analyze the sensitivity of DLDL-v2 to different face regions. We occlude different portions of the input image by setting it to mean values of all training images. Specifically, we use two type of occlusions, small square region (size of 32$\times$32) and horizontal stripe (size of 32$\times$224), as in ~\cite{zeiler2014visualizing, rothe2016deep}. We occlude the input images (size of 224$\times$224) using this two type of occlusions in a sliding window fashion. In all, we obtain 49+7 occluded inputs for each input image. For each occluded input, we record prediction performance (\ie, MAE) on all testing images. Finally, we compute the relative performance loss between with and without occlusions to measure the sensitivity of occlusion region.

\begin{figure}
    \captionsetup[subfigure]{labelformat=empty}
    \captionsetup[subfigure]{justification=centering}
    \centering
    \subfloat[]{\includegraphics[width= 0.15\columnwidth,keepaspectratio]{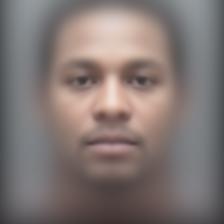}}
    \subfloat[]{\includegraphics[width= 0.15\columnwidth,keepaspectratio]{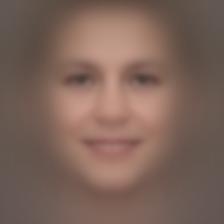}}
    \subfloat[]{\includegraphics[width= 0.15\columnwidth,keepaspectratio]{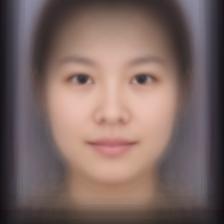}}
    \subfloat[]{\includegraphics[width= 0.15\columnwidth,keepaspectratio]{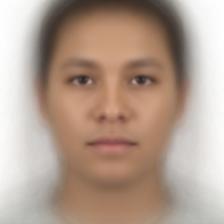}}
    \\ \vspace{-22pt}
    \subfloat[]{\includegraphics[width= 0.15\columnwidth,keepaspectratio]{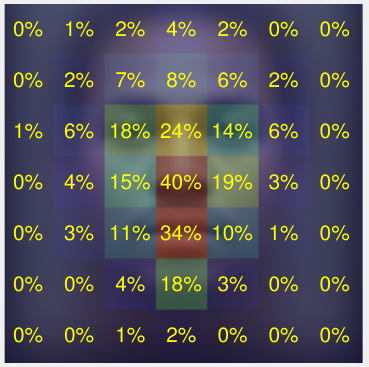}}
    \subfloat[]{\includegraphics[width= 0.15\columnwidth,keepaspectratio]{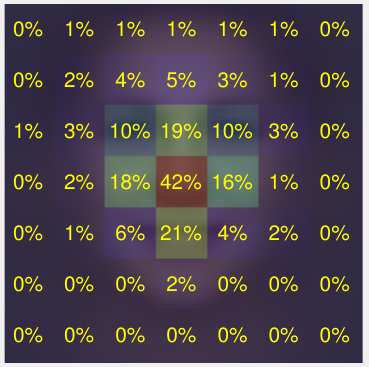}}
    \subfloat[]{\includegraphics[width= 0.15\columnwidth,keepaspectratio]{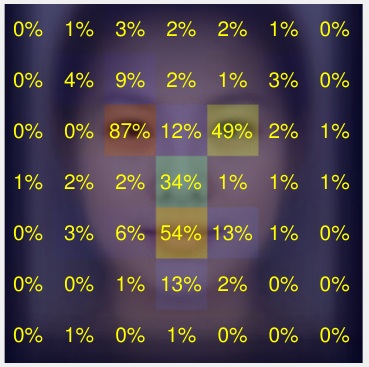}}
    \subfloat[]{\includegraphics[width= 0.15\columnwidth,keepaspectratio]{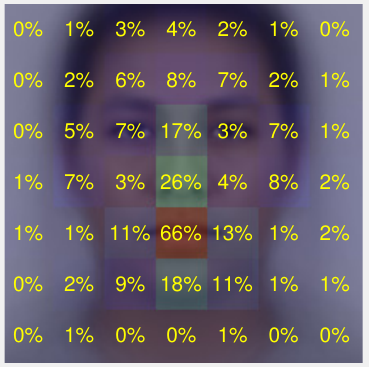}}
    \\ \vspace{-22pt}
    \subfloat[][Morph]{\includegraphics[width= 0.15\columnwidth,keepaspectratio]{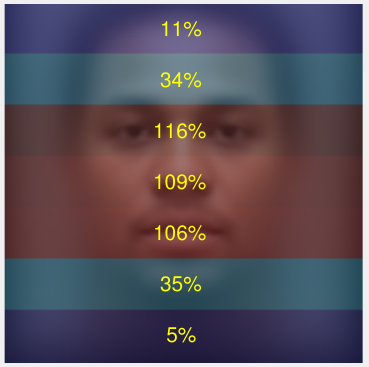}}
    \subfloat[][ChaLearn16]{\includegraphics[width= 0.15\columnwidth,keepaspectratio]{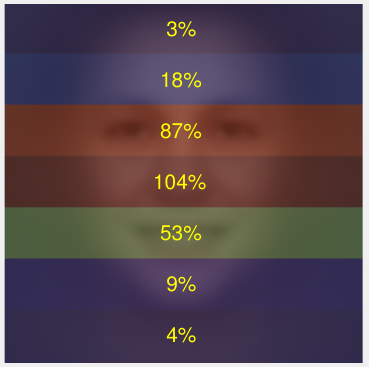}}
    \subfloat[][SCUT-FBP]{\includegraphics[width= 0.15\columnwidth,keepaspectratio]{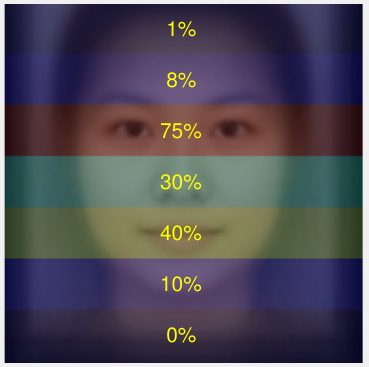}}
    \subfloat[][CFD]{\includegraphics[width= 0.15\columnwidth,keepaspectratio]{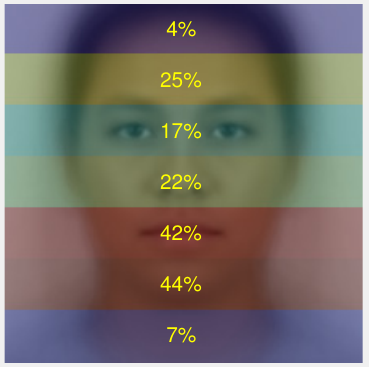}}
 \caption{Sensitivity to different face regions on facial age and attractiveness estimation datasets. The first row shows average image of all testing images on four datasets. The second and third row show heatmaps of the relative performance loss under two occlusions, respectively~(Best viewed in color).}\label{fig:occvi}  
 \vspace{-15pt}
\end{figure}

In Fig.~\ref{fig:occvi}, we show the quantitative results under different occlusions. First, we observe that larger values usually appear in some specific regions such as forehead, two eyes, nose, mouth, and chin. This indicates the decision of DLDL-v2 heavily depends on these crucial regions. Second, these values are significantly different in the different regions. For example, on ChaLearn16 testing images, the largest and second largest value appear around nose and eyes, which suggests nose and eyes are the most important facial traits for age estimation. Third, although SCUT-FBP and CFD both are used to evaluate facial attractiveness datasets, the distributions of the largest value are greatly different. The largest value of the former appears the region of eyes, and that of the latter appears the region of mouth and chin. In fact, the faces of SCUT-FBP come from Asia female, which is scored by Chinese, while CFD dataset consists multi-race faces and is scored by diverse background annotators. Therefore, this difference may be due to the phenomenon that different races may have an inconsistent understanding of facial attractiveness.

\subsection{Why does DLDL-v2 make good estimation?}
Compared to MR, the training procedure of our DLDL-v2 is more stable because it not only regresses the single value with expectation module but also learns a label distribution. Compared to DEX, through introducing label distribution learning module to DLDL-v2, the training instances associated with each class label is significantly increased without actually increasing the number of the total training images, which effectively alleviate the risk of over-fitting. 

For Ranking and DLDL-based methods, we have proved that they are both learning a label distribution from different levels. Therefore, they both share the advantages of label distribution learning. However, there are three major differences in the network architectural between these methods
and our DLDL-v2. First, these methods depend heavily on a pre-trained model such as VGGNet or VGGFace with more parameters while DLDL-v2 has a thinner architecture with fewer parameters. Therefore, DLDL-v2 has higher efficiency in inference time and lower storage overhead. Second, DLDL-v2 effectively avoids the inconsistency between training objective and evaluation metric via introducing the expectation regression module. Third, DLDL-v2 is a fully convolutional network, which removes all but the final fully connected layer. It is very helpful to understand that how DLDL-v2 makes the final decision {because of the parameter sharing mechanism between Eq.~\ref{eq:linear} and~\ref{eq:classmap}.} In a word, these differences make DLDL-v2 have good performance on accuracy, speed, model size and interpretability.

\section{Conclusion}\label{sec:co}
In this paper, we proposed a solution for facial age and attractiveness estimation problems. We firstly analyze that Ranking-based methods are implicitly learning label distribution as DLDL-based methods. This result unifies existing state-of-the-art facial age and attractiveness estimation methods into the DLDL framework. Second, our proposed DLDL-v2 framework can effectively erase the inconsistency between training and evaluation stages via jointly learning label distribution and regressing single value with a thin and deep network architecture. It creates new state-of-the-art results on facial age and attractive estimation tasks with fewer parameters and faster speed, which indicates it is easy to be deployed on resource-constrained devices. In addition, our DLDL-v2 is also a partly interpretable deep framework which employs different patterns to estimate facial attributes.

It is noteworthy that our approach is easily scalable to others label uncertainty tasks, such as skeletal maturity assessment on pediatric hand radiographs~\cite{larson2017performance}, head pose estimation~\cite{schwarz2017driveahead}, popularity of selfie~\cite{kalayeh2015selfie}, image aesthetic assessment~\cite{deng2017image} \etc  ~In addition, a further theoretical study between the ranking-CNN and DLDL will also be our future work.

\bibliography{egbib}

\end{document}